\documentclass[10pt,twocolumn,letterpaper]{article}

\newif\ifsupp
\suppfalse
\def\addsupp{\global\supptrue}

\addsupp 

\ifsupp
\usepackage{titling} 
\usepackage{chapterbib} 
\usepackage[numbers]{natbib} 
\usepackage{setspace} 

\usepackage{xpatch}
  {\begin{titlepage}\def\@thanks{}}%
  {\end{titlepage}}
\xpatchcmd\titlepage{\setcounter{page}\@ne}{}{}{}
\xpatchcmd\endtitlepage{\setcounter{page}\@ne}{}{}{}
\fi

\usepackage{cvpr}
\usepackage{times}
\usepackage{epsfig}
\usepackage{graphicx}
\usepackage{amsmath}
\usepackage{amssymb}
\usepackage{bm}
\usepackage{overpic}
\usepackage[usenames]{color}
\usepackage{color, colortbl}
\usepackage{lipsum}
\usepackage{enumitem} 
\usepackage[ruled,linesnumbered,titlenumbered]{algorithm2e}

\definecolor{DarkGreen}{rgb}{0.0,0.7,0.0} 
\usepackage[pagebackref=true,breaklinks=true,letterpaper=true,colorlinks,bookmarks=false,citecolor=DarkGreen]{hyperref}

\cvprfinalcopy 

\def\cvprPaperID{1556} 

\newcommand{\myparagraph}[1]{\vskip 0.5mm \noindent \textbf{#1}}

\newcommand{\argmin}{\operatornamewithlimits{argmin}}
\makeatletter
\newcommand{\figcaption}[1]{\def\@captype{figure}\caption{#1}}
\newcommand{\tblcaption}[1]{\def\@captype{table}\caption{#1}}
\makeatother
\def\figw{0.49\linewidth}

\def\myD{\mathcal{D}}
\def\myF{\mathcal{F}}
\def\myS{\mathcal{S}}
\def\myO{\mathcal{O}}
\def\myU{\mathcal{U}}
\def\myp{\mathbf{p}}
\def\myq{\mathbf{q}}
\def\myP{\mathbf{P}}
\def\mys{\mathbf{s}}

\ifcvprfinal\fi
\begin{document}
\title{Fast Multi-frame Stereo Scene Flow with Motion Segmentation} 


\ifsupp

\pagenumbering{gobble}
\title{Fast Multi-frame Stereo Scene Flow with Motion Segmentation}

\author{Tatsunori Taniai\thanks{Work done during internship at Microsoft Research and partly at the University of Tokyo. 
}
\\
RIKEN AIP\\
\and
Sudipta N. Sinha\\
Microsoft Research\\
\and
Yoichi Sato\\
The University of Tokyo
}

\maketitle

\begin{abstract}
We propose a new multi-frame method for efficiently computing scene flow (dense depth and optical flow) and camera ego-motion for a dynamic scene observed from a moving stereo camera rig. Our technique also segments out moving objects from the rigid scene.
In our method, we first estimate the disparity map and the 6-DOF camera motion using stereo matching and visual odometry.  We then identify regions inconsistent with the estimated camera motion and compute per-pixel optical flow only at these regions. This flow proposal is fused with the camera motion-based flow proposal using fusion moves to obtain the final optical flow and motion segmentation.
This unified framework benefits all four tasks -- stereo, optical flow, visual odometry and motion segmentation leading to overall higher accuracy and efficiency.
Our method is currently ranked third on the KITTI 2015 scene flow benchmark. Furthermore, our CPU implementation runs in 2-3 seconds per frame which is 1-3 orders of magnitude faster than the top six methods. We also report a thorough evaluation on challenging Sintel sequences with fast camera and object motion, where our method consistently outperforms OSF~\cite{Menze2015}, which is currently ranked second on the KITTI benchmark.
\end{abstract}

\section{Introduction}

Scene flow refers to 3D flow or equivalently the dense 3D motion field of a scene~\cite{Vedula1999}. It can be estimated from video acquired with synchronized cameras from multiple viewpoints~\cite{Lv2016,Mayer2016,Menze2015,Vogel2015} or with RGB-D sensors~\cite{Hornacek2014,Jaimez2015a,Herbst2013,Quiroga2014} and has applications in video analysis and editing, 3D mapping, autonomous driving~\cite{Menze2015} and mobile robotics.

Scene flow estimation builds upon two tasks central to computer vision -- stereo matching and optical flow estimation. Even though many existing methods can already solve these two tasks independently ~\cite{Kolmogorov2001,Hirschmuller2008,Taniai2014,Lucas81,Horn81,Xu2012,Chen2016}, a naive combination of stereo and optical flow methods for computing scene flow is unable to exploit inherent redundancies in the two tasks or leverage additional scene information which may be available.
Specifically, it is well known that the optical flow between consecutive image pairs for stationary (rigid) 3D points are constrained by their depths and the associated 6-DOF motion of the camera rig. However, this idea has not been fully exploited by existing scene flow methods. 
Perhaps, this is due to the additional complexity involved in simultaneously estimating camera motion and detecting moving objects in the scene. 


\begin{figure}[!t]
\centering
\footnotesize
\includegraphics[width=\figw]{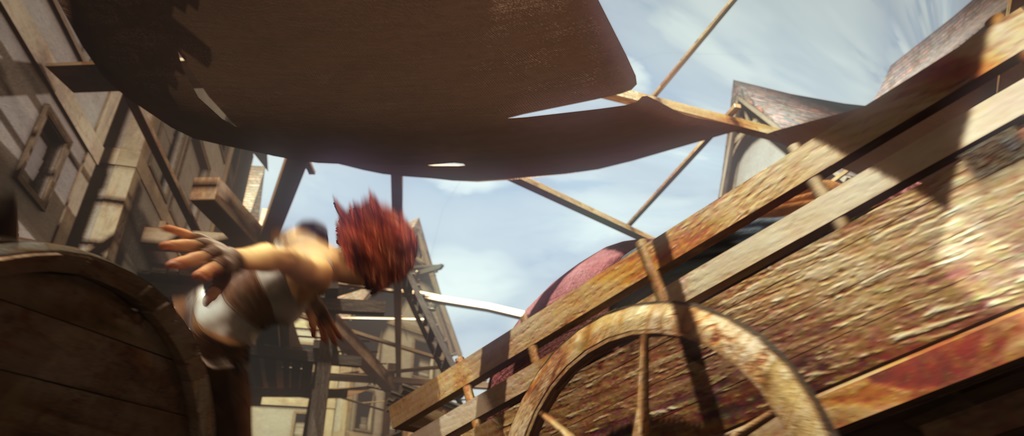}\hfill
\includegraphics[width=\figw]{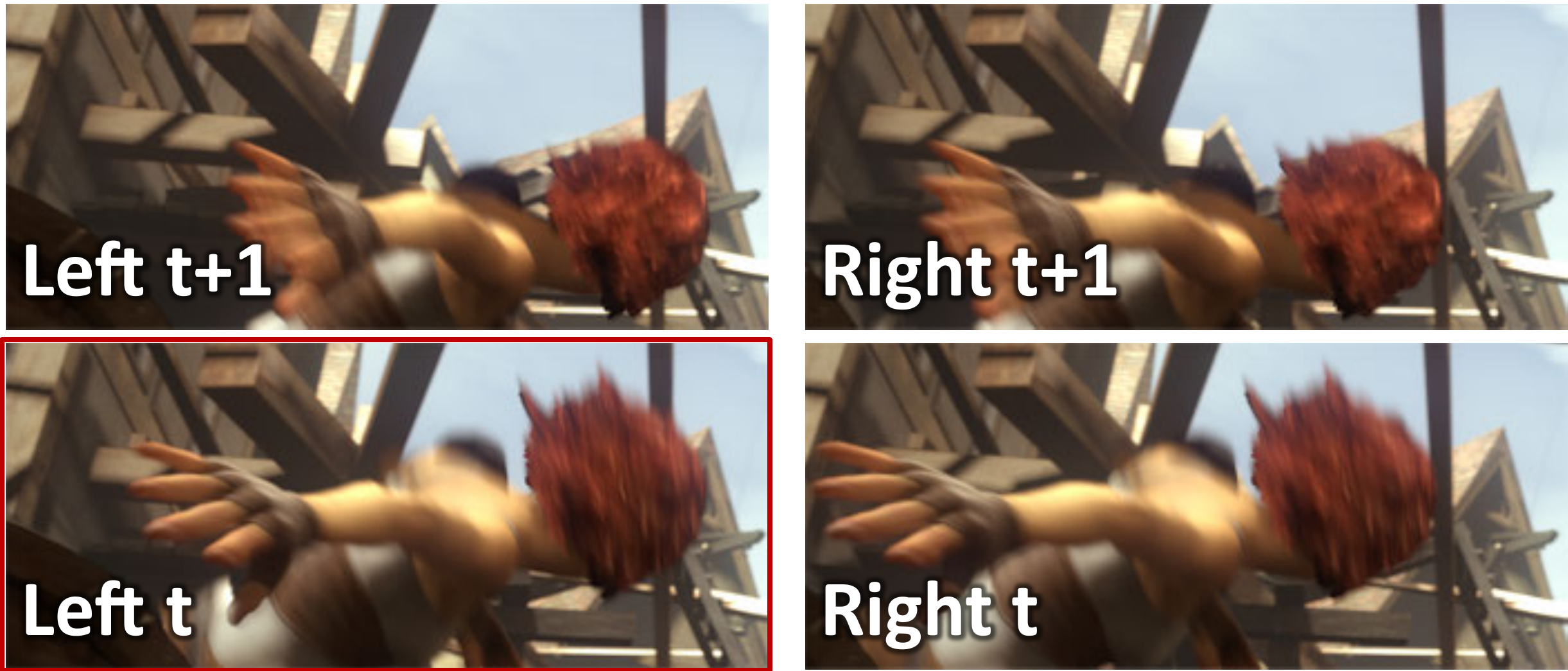}\\
\makebox[\figw]{(a) Left input frame (reference)}\hfill
\makebox[\figw]{(b) Zoom-in on stereo frames}\\
\vskip 0.5mm
\includegraphics[width=\figw]{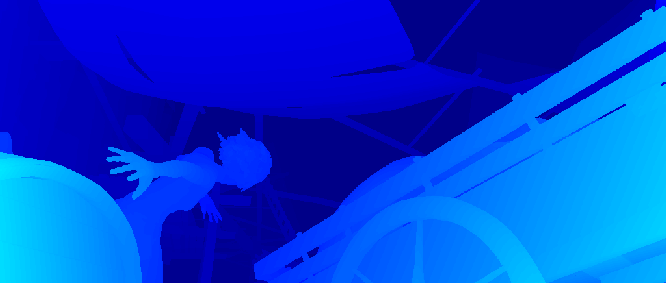}\hfill
\includegraphics[width=\figw]{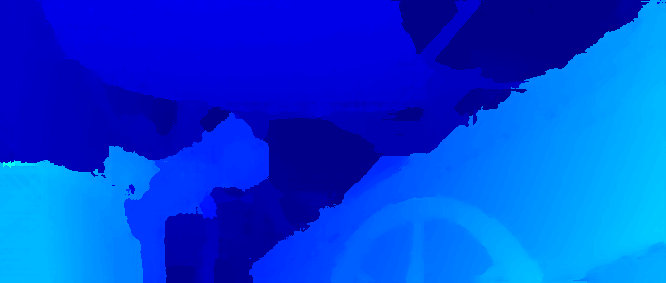}\\
\makebox[\figw]{(c) Ground truth disparity}\hfill
\makebox[\figw]{(d) Estimated disparity $\myD$}\\
\vskip 0.5mm
\includegraphics[width=\figw]{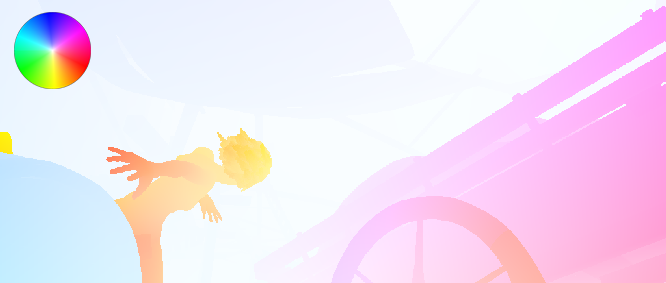}\hfill
\includegraphics[width=\figw]{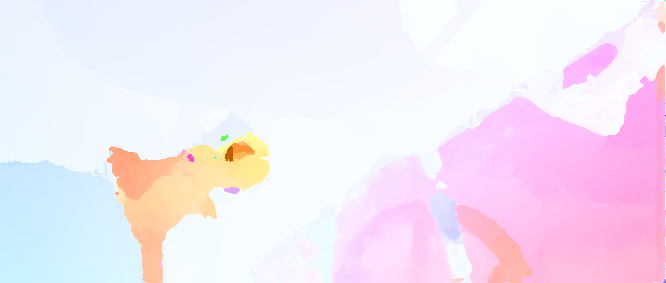}\\
\makebox[\figw]{(e) Ground truth flow}\hfill
\makebox[\figw]{(f) Estimated flow $\myF$}\\
\vskip 0.5mm
\includegraphics[width=\figw]{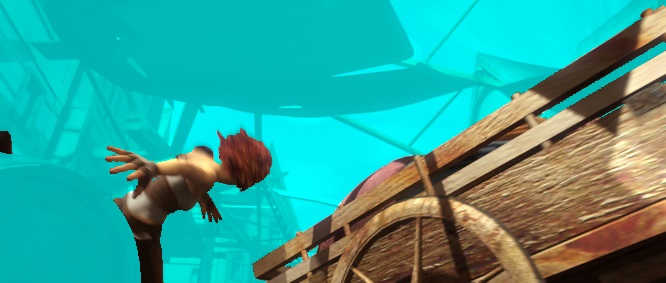}\hfill
\includegraphics[width=\figw]{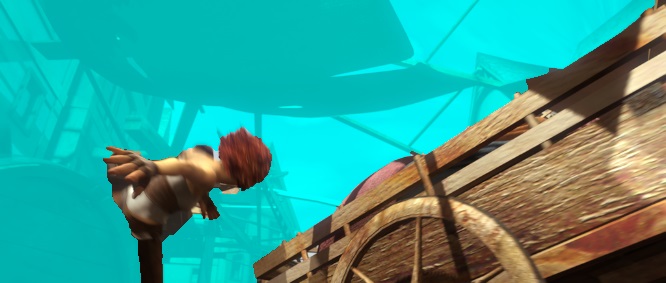}\\
\makebox[\figw]{(g) Ground truth segmentation}\hfill
\makebox[\figw]{(h) Estimated segmentation $\myS$}\\
\vskip 0.5mm
\caption{Our method estimates dense disparity and optical flow from stereo pairs, which is equivalent to stereoscopic scene flow estimation. The camera motion is simultaneously recovered and allows moving objects to be explicitly segmented in our approach.}
\label{fig:example}
\end{figure}

\begin{figure*}[!t]
\centering
\vskip -1mm
\includegraphics[width=0.9\linewidth]{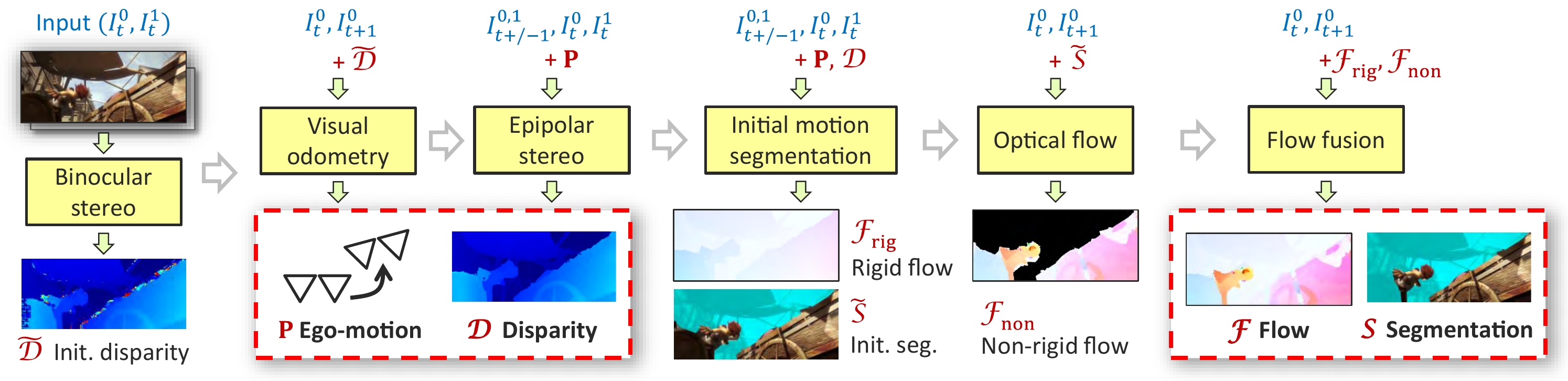}
\caption{Overview of the proposed method. %
In the first three steps, we estimate the disparity $\myD$ and camera motion $\myP$ using stereo matching and visual odometry techniques.
We then detect moving object regions by using the rigid flow $\myF_\text{rig}$ computed from $\myD$ and $\myP$.
Optical flow is performed only for the detected regions, and the resulting non-rigid flow $\myF_\text{non}$ is fused with $\myF_\text{rig}$ to obtain final flow $\myF$ and segmentation~$\myS$.
}
\label{fig:overview}
\end{figure*}

Recent renewed interest in stereoscopic scene flow estimation has led to improved accuracy on challenging benchmarks, which stems from better representations, priors, optimization objectives as well as the use of better optimization methods~\cite{Huguet2007,Wedel2011,Cech2011,Menze2015,Vogel2015,Lv2016}. However, those state of the art methods are computationally expensive which limits their practical usage. In addition, other than a few exceptions~\cite{Vogel2014}, most existing scene flow methods process every two consecutive frames independently and cannot efficiently propagate information across long sequences.

In this paper, we propose a new technique to estimate scene flow from a multi-frame sequence acquired by a calibrated stereo camera on a moving rig. We simultaneously compute dense disparity and optical flow maps on every frame. In addition, the 6-DOF relative camera pose between consecutive frames is estimated along with a per-pixel binary mask that indicates which pixels correspond to either rigid or non-rigid independently moving objects (see Fig.~\ref{fig:example}). Our sequential algorithm uses information only from the past and present,  thus useful for real-time systems.


We exploit the fact that even in dynamic scenes, many observed pixels often correspond to static rigid surfaces. Given disparity maps estimated from stereo images, we robustly compute the 6-DOF camera motion using visual odometry robust to outliers (moving objects in the scene). Given the ego-motion estimate, we improve the depth estimates at occluded pixels via epipolar stereo matching. Then, we identify image regions inconsistent with the camera motion and compute an explicit optical flow proposal for these regions. Finally, this flow proposal is fused with the camera motion-based flow proposal using fusion moves to obtain the final flow map and motion segmentation. 

While these four tasks -- stereo, optical flow, visual odometry and motion segmentation have been extensively studied, most of the existing methods solve these tasks independently. As our primary contribution, we present a single unified framework where the solution to one task benefits the other tasks. 
In contrast to some joint methods~\cite{Vogel2015,Menze2015,Lv2016,Vogel2013} that try to optimize single complex objective functions, we decompose the problem into simpler optimization problems leading to increased computational efficiency.
Our method is significantly faster than top six methods on KITTI taking about 2--3 seconds per frame (on the CPU), whereas state-of-the-art methods take 1--50 minutes per-frame~\cite{Vogel2015,Menze2015,Lv2016,Vogel2013}. Not only is our method faster but it also explicitly recovers the camera motion and motion  segmentation. We now discuss how our unified framework benefits each of the four individual tasks.

\textbf{Optical Flow.} Given known depth and camera motion, the 2D flow for  rigid 3D points which we refer to as {\em rigid flow} in the paper, can be recovered more efficiently and accurately compared to generic {\em non-rigid flow}. 
We still need to compute non-rigid flow but only at pixels associated with moving objects. This reduces redundant computation. 
Furthermore, this representation is effective for occlusion.  Even when corresponding points are invisible in consecutive frames,
the rigid flow can be correctly computed  as long as the depth and camera motion estimates are correct.

\textbf{Stereo.} For rigid surfaces in the scene, our method can recover more accurate disparities at pixels with left-right stereo occlusions. This is because computing camera motions over consecutive frames makes it possible to use multi-view stereo matching on temporally adjacent stereo frames in addition to the current frame pair.

\textbf{Visual Odometry.} Explicit motion segmentation makes camera motion recovery more robust. In our method, the binary mask from the previous frame is used to predict which pixels in the current frame are likely to be outliers and must be downweighted during visual odometry estimation.

\textbf{Motion Segmentation.} This task is essentially solved for free in our method. Since the final optimization performed on each frame fuses rigid and non-rigid optical flow proposals (using MRF fusion moves) the resulting binary labeling indicates which pixels belong to non-rigid objects.


\section{Related Work}
Starting with the seminal work by Vedula~\etal~\cite{Vedula1999,Vedula2005}, the task of estimating scene flow from multiview image sequences has often been formulated as a variational problem~\cite{Pons2003,Pons2007,Basha2010,Wedel2011}. These problems were solved using different optimization methods -- Pons~\etal~\cite{Pons2003,Pons2007} proposed a solution based on level-sets for volumetric representations whereas Basha~\etal~\cite{Basha2010} proposed view-centric representations suiltable for occlusion reasoning and large motions. Previously, Zhang~\etal~\cite{Zhang2001} studied how image segmentation cues can help recover accurate motion and depth discontinuities in multi-view scene flow.

Subsequently, the problem was studied in the binocular stereo setting~\cite{Li2008,Huguet2007,Wedel2011}.
Huguet and Devernay~\cite{Huguet2007} proposed a variational method suitable for the two-view case and Li and Sclaroff~\cite{Li2008} proposed a multiscale approach that incorporated uncertainty during coarse to fine processing. Wedel~\etal~\cite{Wedel2011} proposed an efficient variational method suitable for GPUs where scene flow recovery was decoupled into two subtasks -- disparity and optical flow estimation. 
Valgaerts~\etal~\cite{Valgaerts2010} proposed a variational method that dealt with stereo cameras with unknown extrinsics. 

Earlier works on scene flow were evaluated on sequences from static cameras or cameras moving in relatively simple scenes (see~\cite{Menze2015} for a detailed discussion). Cech~\etal proposed a seed-growing method for sterescopic scene flow~\cite{Cech2011} which could handle realistic scenes with many moving objects captured by a moving stereo camera. The advent of the KITTI benchmark led to further improvements in this field.
Vogel~\etal~\cite{Vogel2011,Vogel2013,Vogel2014,Vogel2015} recently explored a type of 3D regularization -- they proposed a   model of dense depth and 3D motion vector fields in \cite{Vogel2011} and later proposed a piecewise rigid scene model (PRSM) in two~\cite{Vogel2013} and multi-frame settings~\cite{Vogel2014,Vogel2015} that treats scenes as a collection of planar segments undergoing rigid motions. While PRSM~\cite{Vogel2015} is the current top method on KITTI, its joint estimation of 3D geometries, rigid motions and superpixel segmentation using discrete-continuous optimization is fairly complex and  computationally expensive. Lv~\etal~\cite{Lv2016} recently proposed a simplified approach to PRSM using continuous optimization and fixed superpixels (named CSF), which is faster than \cite{Vogel2015} but is still too slow for practical use.

As a closely related approach to ours, object scene flow (OSF)~\cite{Menze2015} segments scenes into multiple rigidly-moving objects based on fixed superpixels, where each object is modeled as a set of planar segments.
This model is more rigidly regularized than PRSM. The inference by max-product particle belief propagation is also very computationally expensive taking 50 minutes per frame. A faster setting of their code takes 2 minutes but has lower accuracy.
A different line of work explored scene flow estimation from RGB-D sequences~\cite{Herbst2013,Quiroga2014,Hornacek2014,Jaimez2015a, Jaimez2015b, Wang2016}. Meanwhile, deep convolutional neural network (CNN) based supervised learning methods have shown promise~\cite{Mayer2016}. 

\section{Notations and Preliminaries}
Before describing our method in details, we define notations and review basic concepts used in the paper.

We denote relative camera motion between two images using matrices $\myP = [\mathbf{R}|\mathbf{t}] \in \mathbb{R}^{3\times 4}$, which transform homogeneous 3D points $\hat{\mathbf{x}} = (x, y, z, 1)^T$ in camera coordinates of the source image to 3D points $\mathbf{x}' = \myP \hat{\mathbf{x}}$ in camera coordinates of the target image.
For simplicity, we assume a rectified calibrated stereo system. Therefore, the two cameras have the same known camera intrinsics matrix $\mathbf{K} \in \mathbb{R}^{3\times 3}$ and the left-to-right camera pose $\myP^{01} = [I|-B\mathbf{e}_x]$ is also known. Here, $I$ is the identity rotation, $\mathbf{e}_x = (1, 0, 0)^T$, and $B$ is the baseline between the left and right cameras.

We assume the  input stereo image pairs have the same size of image domains $\Omega \in \mathbb{Z}^2$ where $\myp=(u,v)^T \in \Omega$ is a pixel coordinate. 
Disparity $\myD$, flow $\myF$ and segmentation $\myS$ are defined as mappings on the image domain $\Omega$, \eg, $\myD(\myp):\Omega \to \mathbb{R}^+$, $\myF(\myp):\Omega \to \mathbb{R}^2$ and $\myS(\myp):\Omega \to \{0, 1\}$.

Given relative camera motion $\myP$ and a disparity map $\myD$ of the source image, pixels $\myp$ of stationary surfaces in the source image are warped to points $\myp' = w(\myp;{\myD},\myP)$ in the target image by the rigid transformation~\cite{Hartley2004} as
\begin{align}
w(\myp;{\myD},\myP) = \pi \left(
\mathbf{K}\mathbf{P} \begin{bmatrix}
\mathbf{K}^{-1} & \mathbf{0}\\
\mathbf{0}^T & (fB)^{-1}\\
\end{bmatrix} \begin{bmatrix}
\hat{\myp}\\
\myD(\myp)\\
\end{bmatrix}
\right). \label{eq:rigidwarp}
\end{align}
Here, $\hat{\myp} = (u, v, 1)^T$ is the 2D homogeneous coordinate of $\myp$, the function $\pi(u,v,w) = (u/w, v/w)^T$ returns 2D non-homogeneous coordinates, 
and $f$ is the focal length of the cameras. 
This warping is also used to find which pixels $\myp$ in the source image are visible in the target image using z-buffering based visibility test and whether \mbox{$\myp' \in \Omega$}.

\section{Proposed Method}
\label{sec:method}
Let $I^0_{t}$ and $I^1_{t}$, $t \in \{1, 2, \cdots, N+1\}$ be the input image sequences captured by the left and right cameras of a calibrated stereo system, respectively.
We sequentially process the first to $N$-th frames and estimate their disparity maps $\mathcal{D}_t$, flow maps $\myF_t$, camera motions $\myP_t$ and motion segmentation masks $\myS_t$ for the left (reference) images. We call moving and stationary objects as foreground and background, respectively.  
Below we focus on processing the \text{$t$-th} frame and omit the subscript $t$ when it is not needed.

At a high level, our method is designed to implicitly minimize image residuals
\begin{equation}
E(\boldsymbol{\Theta}) = \sum_{\myp} \| I^0_t(\myp) - I^0_{t+1}(w(\myp;\boldsymbol{\Theta})) \|
\end{equation}
by estimating the parameters $\boldsymbol{\Theta}$ of the warping function $w$
\begin{equation}
\boldsymbol{\Theta} = \{\mathcal{D}, \myP, \myS, \myF_\text{non} \}.
\end{equation}
The warping function is defined, in the form of the flow map $w(\myp;\mathbf{\Theta}) = \myp + \myF(\myp)$, using the binary segmentation $\mathcal{S}$ on the reference image $I^0_t$ as follows.
\begin{equation}
\myF(\myp) = \left\{
    \begin{array}{ll}
           \myF_\text{rig}(\myp) & \text{if } \myS(\myp) = \text{\textit{background}}\\
       \myF_\text{non}(\myp) & \text{if } \myS(\myp) = \text{\textit{foreground}}
    \end{array}
  \right.
\end{equation}
Here, $\myF_\text{rig}(\myp)$ is the rigid flow computed from the disparity map $\mathcal{D}$ and the camera motion $\myP$ using Eq.~(\ref{eq:rigidwarp}), and $\myF_\text{non}(\myp)$ is the non-rigid flow defined non-parametrically.
Directly estimating this full  model is computationally expensive. 
Instead, we start with a simpler rigid motion model computed from the reduced model parameters $\boldsymbol{\Theta} = \{\mathcal{D}, \myP\}$ (Eq.~(\ref{eq:rigidwarp})), and then increase the complexity of the motion model by adding non-rigid motion regions $\myS$ and their flow $\myF_\text{non}$.
Instead of directly comparing pixel intensities, at various steps of our method, we robustly evaluate the image residuals $\| I(\myp) - I'(\myp')) \|$ by truncated normalized cross-correlation
\begin{equation}
\text{TNCC}_\tau(\myp,\myp') = \min\{ 1 - \text{NCC}(\myp, \myp'), \tau \}. \label{eq:tncc}
\end{equation}
Here, $\text{NCC}$ is normalized cross-correlation computed for $5\times 5$ grayscale image patches centered at $I(\myp)$ and $I'(\myp')$, respectively. The thresholding value $\tau$ is set to 1.

In the following sections, we describe the proposed pipeline of our method.
We first estimate an initial disparity map $\tilde{\mathcal{D}}$ (Sec.~\ref{sec:stereo}).
The disparity map $\tilde{\mathcal{D}}$ is then used to estimate the camera motion $\myP$ using visual odometry recovery (Sec.~\ref{sec:vo}). 
This motion estimate $\myP$ is used in the epipolar stereo matching stage, where we improve the initial disparity to get the final disparity map $\mathcal{D}$ (Sec.~\ref{sec:epistereo}). The $\mathcal{D}$ and $\myP$ estimates are used to compute a rigid flow proposal $\myF_\text{rig}$ and recover an initial segmentation  $\tilde{\myS}$ (Sec.~\ref{sec:segment}). We then estimate non-rigid flow proposal $\myF_\text{non}$ for only the moving object regions of $\tilde{\myS}$ (Sec.~\ref{sec:optflow}). Finally we fuse the rigid and non-rigid flow proposals $\{\myF_\text{rig}, \myF_\text{non}\}$ and obtain the final flow map $\myF$ and segmentation $\myS$ (Sec.~\ref{sec:fusion}).
All the steps of the proposed method are summarized in Fig.~\ref{fig:overview}.

\subsection{Binocular Stereo}
\label{sec:stereo}
\begin{figure}[t]
\footnotesize
\centering
\includegraphics[width=\figw]{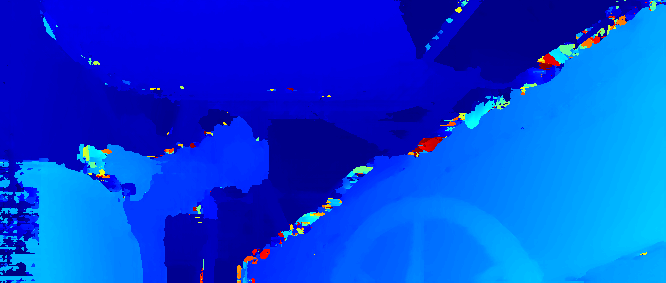}\hfill
\includegraphics[width=\figw]{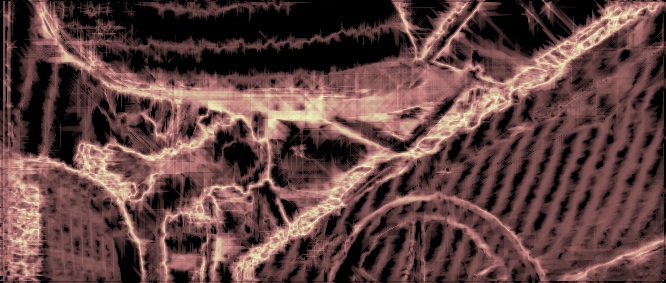}\\
\makebox[\figw]{(a) Initial disparity map $\mathcal{\tilde{D}}$}\hfill
\makebox[\figw]{(b) Uncertainty map $\myU$~\cite{Drory2014}}\\
\vskip 1mm
\includegraphics[width=\figw]{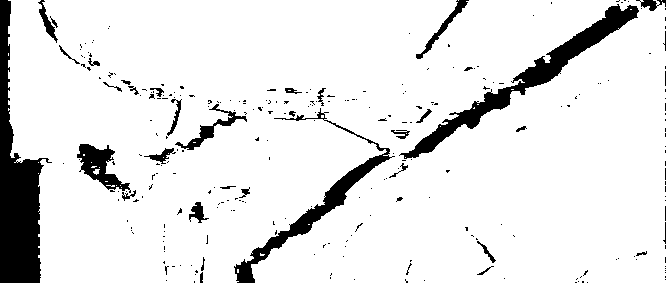}\hfill
\includegraphics[width=\figw]{figures/example2/epidisp0}\\
\makebox[\figw]{(c) Occlusion map $\myO$}\hfill
\makebox[\figw]{(d) Final disparity map $\myD$}\\
\vskip 1mm
\caption{Binocular and epipolar stereo. (a) Initial disparity map. (c) Uncertainity map~\cite{Drory2014} (darker pixels are more confident). (b)~Occlusion map (black pixels are invisible in the right image). (d)~Final disparity estimate by epipolar stereo.}
\label{fig:stereo}
\end{figure}
Given left and right images $I^0$ and $I^1$, we first estimate an initial disparity map $\mathcal{\tilde{D}}$ of the left image and also its occlusion map $\myO$ and uncertainty map $\myU$~\cite{Drory2014}. We visualize example estimates in Figs.~\ref{fig:stereo}~(a)--(c).

As a defacto standard method, we estimate disparity maps by using semi-global matching (SGM)~\cite{Hirschmuller2008} with a fixed disparity range of $[0, 1, \cdots, D_\text{max}]$.
Our implementation of SGM uses 8 cardinal directions and NCC-based matching costs of Eq.~(\ref{eq:tncc}) for the data term.
The occlusion map $\myO$ is obtained by left-right consistency check.
The uncertainty map $\myU$ is computed during SGM as described in \cite{Drory2014} without any computational overhead. We also define a fixed confidence threshold $\tau_\text{u}$ for $\myU$, \ie, $\tilde{\myD}(\myp)$ is considered unreliable if $\myU(\myp) > \tau_\text{u}$.
More details are provided in the supplementary material.

\subsection{Stereo Visual Odometry}
\label{sec:vo}
Given the current and next image $I^0_{t}$ and $I^0_{t+1}$ and the initial disparity map $\tilde{\myD}_t$ of $I^0_{t}$, we estimate the relative camera motion $\myP$ between the current and next frame.
Our method extends an existing stereo visual odometry method~\cite{Alismail2014}. This is a direct method, \ie, it estimates the 6-DOF camera motion $\myP$ by directly minimizing image intensity residuals
\begin{equation}
E_\text{vo}(\myP) = \sum_{\myp \in T} \omega_\myp^\text{vo} \rho \left(| I^0_t(\myp) - I^0_{t+1}(w(\myp;\tilde{\myD}_t,\myP)) | \right)
\end{equation}
for some target pixels $\myp \in T$, using the rigid warping  $w$ of Eq.~(\ref{eq:rigidwarp}).
To achieve robustness to outliers (\eg, by moving objects, occlusion, incorrect disparity), the residuals are scored using the Tukey's bi-weight~\cite{Beaton74}
 function denoted by $\rho$.
The energy $E_\text{vo}$ is minimized by iteratively re-weighted least squares in the inverse compositional framework~\cite{Baker2004}. 

We have modified this method as follows.
First, to exploit motion segmentation available in our method, we adjust the weights $\omega_\myp^\text{vo}$ differently. They are set to either 0 or 1 based on the occlusion map $\myO(\myp)$ but later downweighted by $1/8$, if  $\myp$ is predicted as a moving object point by the previous mask ${\myS}_{t-1}$ and flow $\myF_{t-1}$.
Second, to reduce sensitivity of direct methods to initialization,
we generate multiple diverse initializations for the optimizer and obtain multiple candidate solutions. We then choose the final estimate $\myP$ such that best minimizes weighted NCC-based residuals $E=\sum_{\myp \in \Omega} \omega_\myp^\text{vo} \text{TNCC}_\tau(\myp,w(\myp;\tilde{\myD}_t,\myP))$.
For diverse initializations, we use (a) the identity motion, (b) the previous motion $\myP_{t-1}$, (c) a motion estimate by feature-based correspondences using \cite{Lepetit2009}, and (d) various forward translation motions (about 16 candidates, used only for driving scenes).

  
\subsection{Epipolar Stereo Refinement}
\label{sec:epistereo}
As shown in Fig.~\ref{fig:stereo}~(a), the initial disparity map $\tilde{\myD}$ computed from the current stereo pair $\{I^0_{t}, I^1_{t}\}$ can have errors at pixels occluded in right image.
To address this issue, we use the multi-view epipolar stereo technique on temporarily adjacent six images $\{I^0_{t-1}, I^1_{t-1},I^0_{t}, I^1_{t}, I^0_{t+1}, I^1_{t+1}\}$ and obtain the final disparity map $\myD$ shown in Fig.~\ref{fig:example}~(d). 

From the binocular stereo stage, we already have  computed a matching cost volume of $I^0_{t}$ for $I^1_{t}$, which we denote as $C_\myp(d)$, with some disparity range $d \in [0, D_\text{max}]$.
The goal here is to get a better cost volume $C^\text{epi}_\myp(d)$ as input to SGM, by blending $C_\myp(d)$ with matching costs for each of the four target images $I' \in \{I^0_{t-1}, I^1_{t-1},I^0_{t+1}, I^1_{t+1}\}$.
Since the relative camera poses of the current to next frame $\myP_{t}$ and previous to current frame $\myP_{t-1}$ are already estimated by the visual odometry in Sec.~\ref{sec:vo},
the relative poses from $I^0_{t}$ to each target image can be estimated as $\myP' \in \{\myP_{t-1}^{-1}, \myP^{01}\myP_{t-1}^{-1}, \myP_{t}, \myP^{01}\myP_{t}\}$, respectively. Recall $\myP^{01}$ is the known left-to-right camera pose. 
Then, for each target image $I'$, we compute matching costs $C'_\myp(d)$ by projecting points $({\myp}, d)^T$ in $I^0_{t}$ to its corresponding points in $I'$ using the pose $\myP'$ and the rigid transformation of Eq.~(\ref{eq:rigidwarp}). Since $C'_\myp(d)$ may be unreliable due to moving objects, we here lower the thresholding value $\tau$ of NCC in Eq.~(\ref{eq:tncc}) to $1/4$ for higher robustness.
The four cost volumes are averaged to obtain ${C}^\text{avr}_\myp(d)$.
We also truncate the left-right matching costs $C_\myp(d)$ at $\tau = 1/4$ at  occluded pixels known by $\myO(\myp)$.

Finally, we compute the improved cost volume $C^\text{epi}_\myp(d)$ by linearly blending  $C_\myp(d)$ with ${C}^\text{avr}_\myp(d)$ as 
\begin{align}
C^\text{epi}_\myp(d) = (1-\alpha_\myp) {C}_\myp(d) + \alpha_\myp {C}^\text{avr}_\myp(d),
\end{align}
and run SGM with $C^\text{epi}_\myp(d)$ to get the final disparity map $\myD$.
The blending weights $\alpha_\myp \in [0, 1]$ are computed from the uncertainty map $\myU(\myp)$ (from  Sec.~\ref{sec:stereo}) normalized as $u_\myp = \min \{ \myU(\myp)/\tau_u, 1\}$ and then converted as follows.
\begin{equation}
\alpha_\myp(u_\myp) = \max \{u_\myp- \tau_\text{c}, 0 \} /(1-\tau_\text{c}).
\end{equation}
Here, $\tau_\text{c}$ is a confidence threshold. 
If $u_\myp \le \tau_\text{c}$, we get $\alpha_\myp = 0$ and thus $C^\text{epi}_\myp = {C}_\myp$. When $u_\myp$ increases from $\tau_\text{c}$ to 1,  $\alpha_\myp$ linearly increases from 0 to 1.
Therefore, we only need to compute ${C}^\text{avr}_\myp(d)$ at $\myp$ where $u_\myp > \tau_\text{c}$, which saves computation.
 We use $\tau_c = 0.1$.

\subsection{Initial Segmentation}
\label{sec:segment}
\begin{figure}[t]
\footnotesize
\centering
\includegraphics[width=\figw]{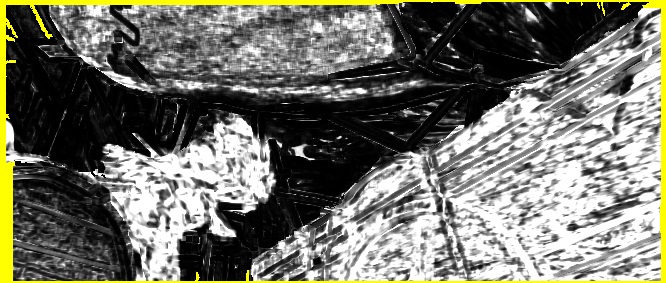}\hfill
\includegraphics[width=\figw]{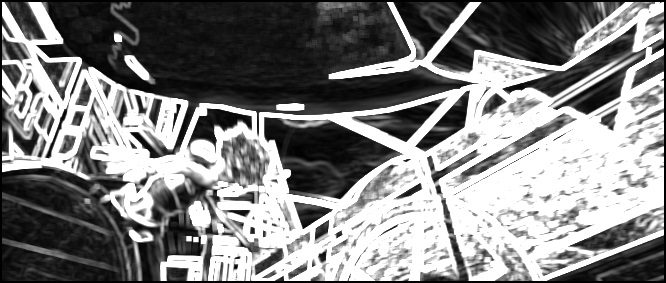}\\
\makebox[\figw]{(a) NCC-based residual map}\hfill
\makebox[\figw]{(b) Patch-intensity variance $w_\myp^\text{var}$}\\
\vskip 0.5mm
\includegraphics[width=\figw]{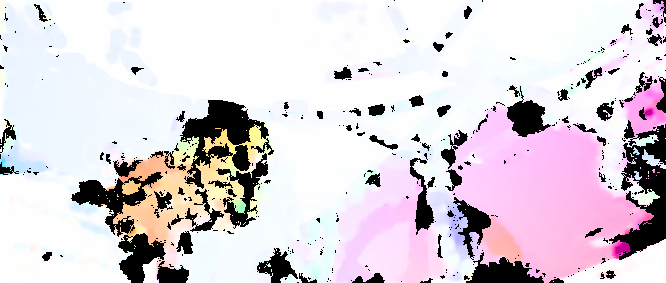}\hfill
\includegraphics[width=\figw]{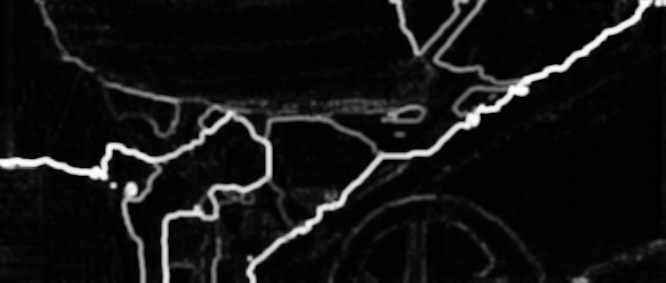}\\
\makebox[\figw]{(c) Prior flow $\myF_\text{pri}$~\cite{Farneback2003}}\hfill
\makebox[\figw]{(d) Depth edge map~$w_{\myp \myq}^\text{dep}$}\\
\vskip 0.5mm
\includegraphics[width=\figw]{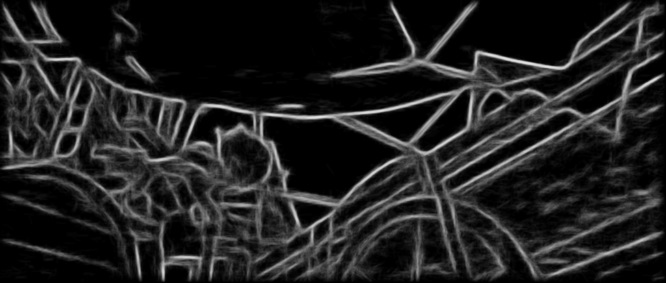}\hfill
\includegraphics[width=\figw]{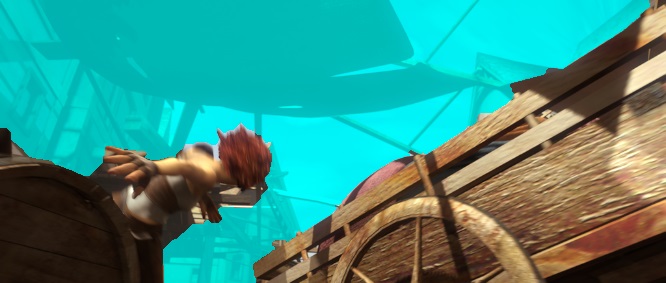}\\
\makebox[\figw]{(e) Image edge map $w_{\myp \myq}^\text{str}$~\cite{Dollar2015}}\hfill
\makebox[\figw]{(f)  Initial segmentation $\tilde{S}$}\\
\vskip 0.5mm
\caption{Initial segmentation.
We detect moving object regions using clues from (a) image residuals weighted by (b) patch-intensity variance and (c) prior flow.
We also use (d) depth edge and (e) image edge information to obtain (f) initial segmentation.
}
\label{fig:segment}
\end{figure}

During the initial segmentation step, the goal is to find a binary segmentation $\tilde{\myS}$ in the reference image $I_t^0$, which shows where the rigid flow proposal $\myF_\text{rig}$ is inaccurate and hence optical flow must be recomputed. Recall that $\myF_\text{rig}$ is obtained from the estimated disparity map $\myD$ and camera motion $\myP$ using Eq.~(\ref{eq:rigidwarp}).
An example of $\tilde{\myS}$ is shown in Fig.~\ref{fig:segment}~(f). We now present the details.

First, we define binary variables $s_\myp \in \{0, 1\}$ as proxy of $\tilde{\myS}(\myp)$ where $1$ and $0$ correspond to foreground (moving objects) and background, respectively. Our segmentation energy $E_\text{seg}(\mys)$ is defined as 
\begin{equation}
E_\text{seg} = \sum_{\myp \in \Omega} \big[  C_\myp^\text{ncc} + C_\myp^\text{flo} + C_\myp^\text{col} +
C_\myp^\text{pri} \big] \bar{s}_\myp + E_\text{potts}(\mys). \label{eq:seg}
\end{equation}
Here, $\bar{s}_\myp = 1 - s_\myp$.
The bracketed terms $\left[ \, \cdot \, \right]$ are data terms that encode the likelihoods for mask $\tilde{\myS}$, \ie, positive values bias $s_\myp$ toward 1 (moving foreground). $E_\text{potts}$ is the pairwise smoothness term. We explain each term below.

\myparagraph{Appearance term $\mathbf{C}_\myp^\text{ncc}$}: This term finds moving objects by checking image residuals of rigidly aligned images.
We compute NCC-based matching costs between  $I = I_t^0$ and $I ' = I_{t+1}^0$ as 
\begin{equation}
{C'}_\myp^\text{ncc}(I,I') =   \text{TNCC}_\tau(\myp,\myp';I,I') - \tau_\text{ncc}
\end{equation}
where $\myp' = \myp + \myF_\text{rig}(\myp)$ and $\tau_\text{ncc} \in (0, \tau)$ is a threshold. 
However, TNCC values are unreliable at texture-less regions (see the high-residual tarp in Fig.~\ref{fig:segment}~(a)).  Furthermore, if $\myp'$ is out of field-of-view, ${C'}_\myp^\text{ncc}$ is not determined (yellow pixels in Fig.~\ref{fig:segment}~(a)). Thus, similarly to  epipolar stereo, we match $I_t^0$ with $I' \in \{I^0_{t-1}, I^1_{t-1},I^0_{t+1}, I^1_{t+1}\}$ and compute the average of valid matching costs
\begin{equation}
{C}_\myp^\text{ncc} = \lambda_\text{ncc} w_\myp^\text{var} \, \text{Average}_{I'} \big[  {C'}_\myp^\text{ncc}(I,I') \big]. \label{eq:seg_ncc}
\end{equation}
Matching with many images increases the recall for detecting moving objects.
To improve matching reliability,  ${C}_\myp^\text{ncc}$ is weighted by $w_\myp^\text{var} = \min(\text{StdDev}(I), \tau_w) / \tau_w$, the truncated standard deviation of the $5 \times 5$ patch centered at $I(\myp)$. The weight map $w_\myp^\text{var}$ is visualized in Fig.~\ref{fig:segment}~(b). We also truncate ${C'}_\myp^\text{ncc}(I,I')$ at 0, if $\myp'$ is expected to be occluded in $I'$ by visibility test. We use $(\lambda_\text{ncc}, \tau_\text{ncc}, \tau_w) = (4, 0.5, 0.005)$.

\myparagraph{Flow term $\mathbf{C}_\myp^\text{flo}$}: This term evaluates flow residuals $r_\myp = \| \myF_\text{rig}(\myp) - \myF_\text{pri}(\myp) \|$ between the rigid flow $\myF$ and (non-rigid) prior flow  $\myF_\text{pri}$ computed by \cite{Farneback2003} (see  Fig.~\ref{fig:segment}~(c)).
Using a threshold $\tau_\myp^\text{flo}$ and the patch-variance weight $w_\myp^\text{var}$, we define ${C}_\myp^\text{flo}$ as
\begin{equation}
{C}_\myp^\text{flo} =  \lambda_\text{flo} w_\myp^\text{var} \, \big[ \min( r_\myp, 2 \tau_\myp^\text{flo}) - \tau_\myp^\text{flo} \big] / \tau_\myp^\text{flo}.  \label{eq:seg_flo}
\end{equation}
The part after $w_\myp^\text{var}$ normalizes $(r_\myp - \tau_\myp^\text{flo})$ to lie within $[-1, 1]$.
The  threshold $\tau_\myp^\text{flo}$ is computed at each pixel $\myp$ by
\begin{equation}
\tau_\myp^\text{flo} = \max(\tau^\text{flo}, \gamma \| \myF_\text{rig}(\myp) \|).
\end{equation}
This way the threshold is relaxed if the rigid motion $\myF_\text{rig}(\myp)$ is large. 
If prior flow $\myF_\text{pri}(\myp)$ is invalidated by bi-directional consistency check (black holes in Fig.~\ref{fig:segment}~(c)), ${C}_\myp^\text{flo}$ is set to 0.
We use $(\lambda_\text{flo}, \tau^\text{flo}, \gamma) = (4, 0.75, 0.3)$. 

\myparagraph{Prior term $\mathbf{C}_\myp^\text{pri}$}: This term encodes segmentation priors based on results from previous frames or on scene context via ground plane detection. Sec.~\ref{sec:details} for the details.

\myparagraph{Color term $\mathbf{C}_\myp^\text{col}$}: This is a standard color-likelihood term~\cite{Boykov2001} for  RGB color vectors $\mathbf{I}_\myp$ of pixels in the reference image $I_t^0(\myp)$:
\begin{equation}
C_\myp^\text{col} = \lambda_\text{col} \Big[ \log \theta_1(\mathbf{I}_\myp) -  \log \theta_0(\mathbf{I}_\myp)\Big]. \label{eq:color}
\end{equation}
We use $\lambda_\text{col} = 0.5$ and $64^3$ bins of histograms for the color models $\{\theta_0, \theta_1\}$.

\myparagraph{Smoothness term $\mathbf{E_\text{potts}}$}: This term is based on the Potts model defined for all pairs of neighboring pixels $(\myp,\myq)\in N$ on the 8-connected pixel grid.
\begin{align}
E_\text{potts}(\mys) = \lambda_\text{potts} \!\!\!\!  \sum_{(\myp,\myq)  \in N} \!\!\!\! ( \omega^\text{col}_{\myp\myq} + \omega^\text{dep}_{\myp\myq} +\omega^\text{str}_{\myp\myq} ) |s_\myp - s_\myq|. \label{eq:potts}
\end{align}
We use three types of edge weights. 
The color-based weight $\omega^\text{col}_{\myp\myq}$ is computed as $\omega_{\myp\myq} ^\text{col}= e^{- \| \mathbf{I}_\myp - \mathbf{I}_\myq \|^2_2/\kappa_1}$ where $\kappa_1$ is estimated as the expected value of $2\| \mathbf{I}_\myp - \mathbf{I}_\myq \|^2_2$ over $(\myp, \myq)\in N$~\cite{Rother2004}.
The depth-based weight $\omega^\text{dep}_{\myp\myq}$ is computed as $\omega^\text{dep}_{\myp\myq} = e^{-|L_\myp + L_\myq| / \kappa_2}$ where $L_\myp = |\Delta D(\myp)|$ is the absolute Laplacian of the disparity map $\myD$. The $\kappa_2$ is estimated similarly to $\kappa_1$.
The edge-based weight $\omega^\text{str}_{\myp\myq}$ uses an edge map $e_\myp \in [0, 1]$ obtained by a fast edge detector~\cite{Dollar2015} and is computed as $\omega^\text{str}_{\myp\myq} = e^{-|e_\myp + e_\myq| / \kappa_3}$.
Edge maps of $\omega^\text{dep}_{\myp\myq}$ and $\omega^\text{str}_{\myp\myq}$ (in the form of $1-w_{\myp\myq}$) are visualized in Figs.~\ref{fig:segment}~(d) and (e).
We use $(\lambda_\text{potts}, \kappa_3) = (10, 0.2)$.


The minimization of $E_\text{seg}(\mys)$ is similar to the GrabCut~\cite{Rother2004} algorithm, \ie,
we alternate between minimizing $E_\text{seg}(\mys)$ using graph cuts~\cite{Boykov2004} and updating the color models $\{\theta_1, \theta_0\}$ of  $C_\myp^\text{col}$ from segmentation $\mys$. 
We run up to five iterations until convergence using dynamic max-flow~\cite{Kohli2007}.


\subsection{Optical Flow}
\label{sec:optflow}
\begin{figure}
\footnotesize
\includegraphics[width=\figw]{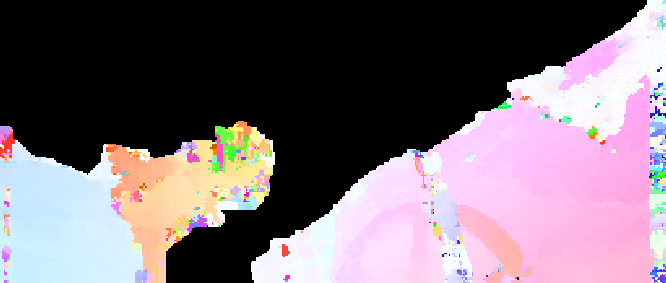}\hfill
\includegraphics[width=\figw]{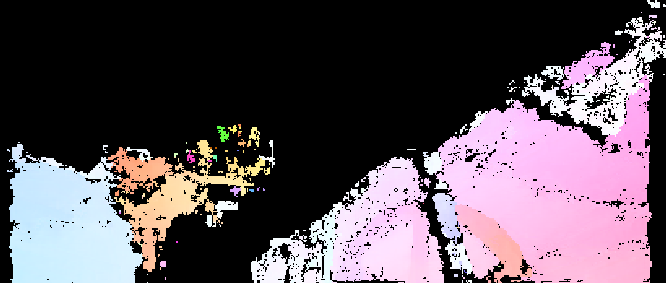}\\
\makebox[\figw]{(a) Non-rigid flow by SGM flow}\hfill
\makebox[\figw]{(b) Consistency check}
\vskip 0.5mm
\includegraphics[width=\figw]{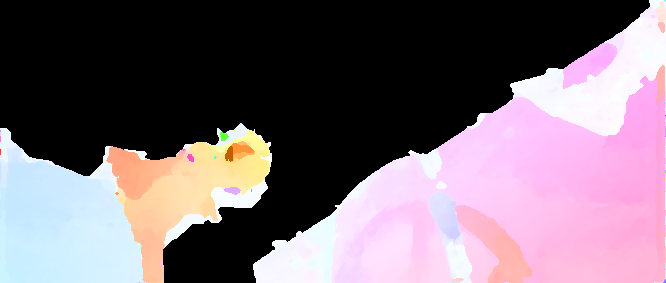}\hfill
\includegraphics[width=\figw]{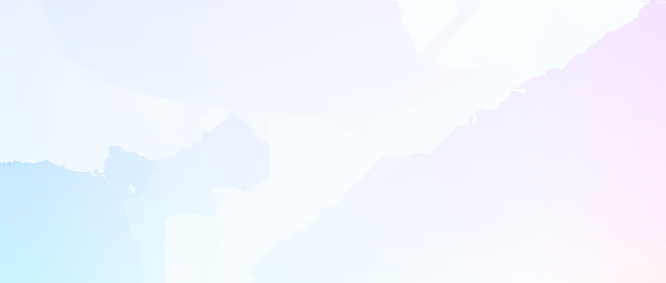}\\
\makebox[\figw]{(c) Non-rigid flow proposal $\myF_\text{non}$}\hfill
\makebox[\figw]{(d) Rigid flow proposal $\myF_\text{rig}$}
\vskip 0.5mm
\includegraphics[width=\figw]{figures/example2/fullflo0}\hfill
\includegraphics[width=\figw]{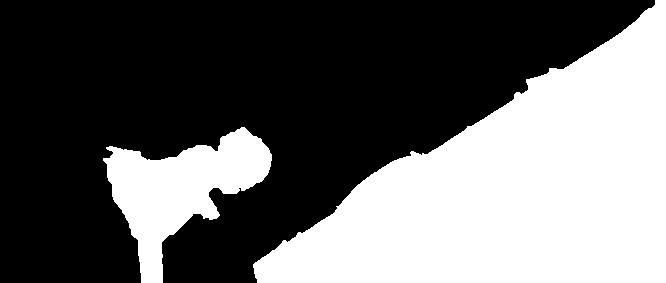}\\
\makebox[\figw]{(e) Final flow map $\myF$}\hfill
\makebox[\figw]{(f) Final segmentation mask  $\myS$}
\vskip 0.5mm
\caption{Optical flow and flow fusion.
We obtain non-rigid flow proposal by (a) performing SGM followed by (b) consistency filtering and (c) hole filing by weighted median filtering. This flow proposal is fused with (d) the rigid flow proposal to obtain (e) the final flow estimate and (f) motion segmentation.
}
\label{fig:optflow}
\end{figure}

Next, we estimate the non-rigid flow proposal $\myF_\text{non}$ for the moving foreground regions estimated as the initial segmentation $\tilde{\myS}$.
Similar to Full Flow~\cite{Chen2016}, we pose optical flow as a discrete labeling problem where the labels represent 2D translational shifts with in a 2D search range (see Sec.~\ref{sec:details} for range estimation).
Instead of \mbox{TRW-S}~\cite{Kolmogorov2006} as used in \cite{Chen2016}, we apply the SGM algorithm as a discrete optimizer. After obtaining a flow map from SGM as shown in Fig.~\ref{fig:optflow}~(a), we filter it further by 1)~doing  bi-directional consistency check (see Fig.~\ref{fig:optflow}~(b)), and 2)~filing holes by weighted median filtering to get the non-rigid flow proposal $\myF_\text{non}$. 
The flow consistency map $\myO^\text{flo}(\myp)$ is passed to the next stage.
Our extension of SGM is straightforward and is detailed in our supplementary material as well as the refinement scheme.

\subsection{Flow Fusion and Final Segmentation}
\label{sec:fusion}
Given the rigid and non-rigid flow proposals $\myF_\text{rig}$ and $\myF_\text{non}$, we fuse them to obtain the final flow estimate $\myF$. This fusion step also produces the final segmentation $\myS$. These inputs and outputs are illustrated in Figs.~\ref{fig:optflow}~(c)--(f).

The fusion process is similar to the initial segmentation. The binary variables $s_\myp\in\{0, 1\}$ indicating the final segmentation $\myS$, now also indicate which of the two flow proposals $\{\myF_\text{rig}(\myp), \myF_\text{non}(\myp)\}$  is selected as the final flow estimate $\myF(\myp)$.
To this end, the energy $E_\text{seg}$ of Eq.~(\ref{eq:seg}) is modified as follows. First,  $C_\myp^\text{ncc}$ is replaced by
\begin{equation}
{C}_\myp^\text{ncc}\! =\! \lambda_\text{ncc} w_\myp^\text{var} [ \text{TNCC}_\tau(\myp,\myp'_\text{rig}) \!-\! \text{TNCC}_\tau(\myp,\myp'_\text{non}) ],
\end{equation}
where $\myp'_\text{rig}=  \myp + \myF_\text{rig}(\myp)$ and $\myp'_\text{non}= \myp + \myF_\text{non}(\myp)$. Second, the prior flow  $\myF_\text{pri}(\myp)$ in $C_\myp^\text{flo}$ is replaced by $\myF_\text{non}(\myp)$.
When $\myp'_\text{rig}$ is out of view or $\myF_\text{non}(\myp)$ is invalidated by the flow occlusion map $\myO^\text{flo}(\myp)$, we set $C_\myp^\text{ncc}$ and $C_\myp^\text{flo}$ to 0.

The fusion step only infers $s_\myp$ for pixels labeled foreground in the initial segmentation $\tilde{\myS}$, since the background labels are fixed. The graph cut optimization for fusion is typically very efficient, since the pixels labeled foreground in $\tilde{\myS}$ is often a small fraction of all the pixels.

\subsection{Implementation Details}
\label{sec:details}
\myparagraph{Disparity range reduction.} For improving the efficiency of epipolar stereo, the disparity range $[0, D_\text{max}]$ is reduced by estimating $D_\text{max}$ from the initially estimated $\tilde{\myD}(\myp)$.
We compute $D_\text{max}$ robustly by making histograms of non-occluded disparities of $\tilde{\myD}(\myp)$ and ignoring bins whose frequency is less than $0.5\%$.  $D_\text{max}$ is then chosen as the max bin from remaining valid non-zero bins.


\myparagraph{Flow range estimation.}
The 2D search range $R=([u_\text{min}, u_\text{max}] \times [v_\text{min}, v_\text{max}])$ for SGM flow  is estimated as follows.
For the target region $\tilde{\myS}$, we compute three such ranges from  feature-based sparse correspondences, 
the prior flow and rigid flow.
For the latter two, we robustly compute ranges by making 2D histograms of flow vectors and ignoring bins whose frequency is less than one-tenth of the max frequency. Then, the final range $R$ is the range that covers all three.
To make $R$ more compact, we repeat the range estimation and subsequent SGM  for individual connected components in $\tilde{\myS}$.

\myparagraph{Cost-map smoothing.} Since NCC and flow-based cost maps $C_\myp^\text{ncc}$ and $C_\myp^\text{flo}$ used in the segmentation and fusion steps are noisy, we smooth them by averaging the values within superpixels. We use superpixelization of approximately 850 segments produced by \cite{VandenBergh2012} in OpenCV. 

\myparagraph{Segmentation priors.} We define $C_\myp^\text{pri}$ of Eq.~(\ref{eq:seg}) as $C_\myp^\text{pri} = \lambda_\text{mask} C_\myp^\text{mask} + C_\myp^\text{pcol}$. Here, $C_\myp^\text{mask} \in [-0.1, 1]$ is a signed soft mask predicted by previous mask $\myS_{t-1}$ and flow $\myF_{t-1}$. Negative background regions are downweighted by 0.1 for better detection of new emerging objects. We use $\lambda_\text{mask}=2$. 
$C_\myp^\text{pcol}$ is a color term similar to Eq.~(\ref{eq:color}) with the same $\lambda_\text{col}$ but uses color models updated online as the average of past color models.
For road scenes, we additionally use the ground prior such as shown in Fig.~\ref{fig:segprior} as a cue for the background. It is derived by the ground plane detected using RANSAC. See the supplementary material for more details.

\begin{figure}[h]
\centering
\vskip -2mm
\footnotesize
\includegraphics[width=0.325\linewidth]{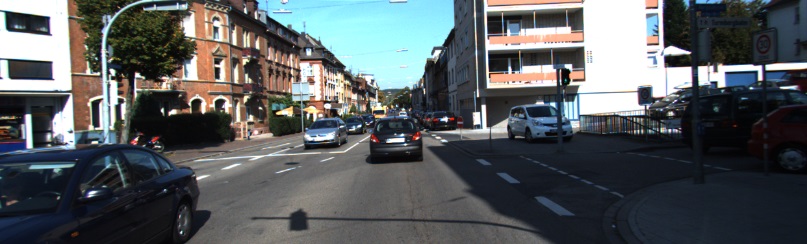}\hfill
\includegraphics[width=0.325\linewidth]{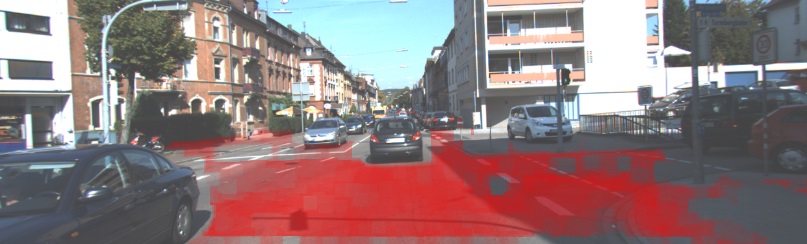}\hfill
\includegraphics[width=0.325\linewidth]{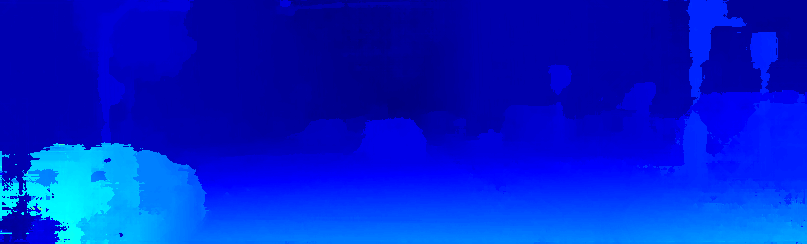}
\caption{Segmentation ground prior. For road scenes (left), we compute the ground prior (middle) from the disparity map (right).}
\label{fig:segprior}
\vskip -2.5mm
\end{figure}

\myparagraph{Others.} We run our algorithm on images downscaled by a factor of $0.4$ for optical flow and $0.65$ for the other steps (each image in KITTI is $1242\times 375$ pixels).
We do a subpixel refinement of the SGM disparity and flow maps via standard local quadratic curve fitting~\cite{Hirschmuller2008}.
\section{Experiments}
We evaluate our method on the KITTI 2015 scene flow benchmark~\cite{Menze2015} and further extensively evaluate on the challenging Sintel (stereo) datasets~\cite{Butler2012}. 
On Sintel we compare with the top two state of the art methods -- PRSM~\cite{Vogel2015} and OSF~\cite{Menze2015}. PRSM is a multi-frame method like ours.
Although OSF does not explicitly distinguish moving objects from static background in segmentation,  the dominant rigid motion bodies are assigned the first object index, which we regarded as background in evaluations.
Our method was implemented in C++ and running times were measured on a computer with a quadcore 3.5GHz CPU.
All parameter settings were determined using KITTI training data for validation.
Only two parameters were re-tuned for Sintel.



\subsection{KITTI 2015 Scene Flow Benchmark}
We show a selected ranking of KITTI benchmark results in Table~\ref{tab:kitti_scores}, where our method is ranked third. Our method is much faster than all the top methods and more accurate than the fast methods~\cite{Derome2016,Cech2011}.
See Fig.~\ref{fig:running_time} for the per-stage running times. The timings for most stages of our method are small and constant, while for optical flow they vary depending on the size of the moving objects.
Motion segmentation results are visually quite accurate (see Fig.~\ref{fig:kitti_results}).
As shown in Table~\ref{tab:epipolar}, epipolar stereo refinement using temporarily adjacent stereo frames improves disparity accuracy even for non-occluded pixels.
By visual inspection of successive images aligned via the camera motion and depth, we verified that there was never any failure in ego-motion estimation.



\subsection{Evaluation on Sintel Dataset}
Unlike previous scene flow methods, we also evaluated our method on Sintel and compared it with OSF~\cite{Menze2015} and PRSM~\cite{Vogel2015} (see Table~\ref{tab:sintel_scores} -- best viewed in color).
Recall, PRSM does not perform motion segmentation.
Although OSF and PRSM are more accurate on KITTI, our method outperforms OSF on Sintel on all metrics. 
Also, unlike OSF, our method is multi-frame. 
Sintel scenes have fast, unpredictable camera motion, drastic non-rigid object motion and deformation unlike KITTI where vehicles are the only type of moving objects.
While OSF and PRSM need strong rigid regularization, we employ per-pixel inference without requiring piecewise planar assumption.
Therefore, our method generalizes more easily to Sintel.
Only two parameters had to be modified as follows. $( \lambda_\text{col}, \tau_\text{ncc} ) = (1.5, 0.25)$.



\myparagraph{Limitations.}
The visual odometry step may fail when the scene is far away (see \textsl{mountain\_1} in  Fig.~\ref{fig:sintel_results}) due to subtle disparity.
It may also fail when the moving objects dominate the field of view.
 Our motion segmentation results are often accurate but in the future we will improve  temporal consistency to produce more coherent motion segmentation.

\begin{table*}[p]
\begin{minipage}{\linewidth}
\definecolor{yellow}{cmyk}{0,0,0.4,0}
\caption{KITTI 2015 scene flow benchmark results~\cite{Menze2015}.
We show the error rates (\%) for the disparity on the reference frame (D1) and second frame (D2), the optical flow (Fl) and the scene flow (SF) at background (bg), foreground (fg) and all pixels.
Disparity or flow is considered correctly estimated if the end-point error is $<$ 3px or $<$ 5\%.
Scene flow is considered correct if D1, D2 and Fl are correct.
}
\label{tab:kitti_scores}
\scriptsize
\centering
\vskip 1mm
\begin{tabular}{c | c | c | c | c | c | c | c | c | c | c | c | c | c | c }
{\bf Rank} & {\bf Method} & {\bf D1-bg} & {\bf D1-fg} & {\bf D1-all} & {\bf D2-bg} & {\bf D2-fg} & {\bf D2-all} & {\bf Fl-bg} & {\bf Fl-fg} & {\bf Fl-all} & {\bf SF-bg} & {\bf SF-fg} & {\bf \textcolor{red}{SF-all}}  & {\bf Time} \\ \hline
1 & PRSM~\cite{Vogel2015} & \textbf{3.02} \ & \textbf{10.52} \ & \textbf{4.27} \ & \textbf{5.13} \ & \textbf{15.11} \ & \textbf{6.79} \ & \textbf{5.33} \ & \textbf{17.02} \ & \textbf{7.28} \ & \textbf{6.61} \ & \textbf{23.60} \ & \textbf{9.44} \  & 300 s \\
2 & OSF~\cite{Menze2015} & 4.54 \ & 12.03 \ & 5.79 \ & 5.45 \ & 19.41 \ & 7.77 \ & 5.62 \ & 22.17 \ & 8.37 \ & 7.01 \ & 28.76 \ & 10.63 \  & 50 min \\
\rowcolor{yellow}
3 & \textbf{FSF+MS} (ours) & 5.72 \ & 11.84 \ & 6.74 \ & 7.57 \ & 21.28 \ & 9.85 \ & 8.48 \ & 29.62 \ & 12.00 \ & 11.17 \ & 37.40 \ & 15.54 \  & 2.7 s \\
4 & CSF~\cite{Lv2016} & 4.57 \ & 13.04 \ & 5.98 \ & 7.92 \ & 20.76 \ & 10.06 \ & 10.40 \ & 30.33 \ & 13.71 \ & 12.21 \ & 36.97 \ & 16.33 \  & 80 s \\
5 & PR-Sceneflow~\cite{Vogel2013} & 4.74 \ & 13.74 \ & 6.24 \ & 11.14 \ & 20.47 \ & 12.69 \ & 11.73 \ & 27.73 \ & 14.39 \ & 13.49 \ & 33.72 \ & 16.85 \  & 150 s \\
8 & PCOF + ACTF~\cite{Derome2016} & 6.31 \ & 19.24 \ & 8.46 \ & 19.15 \ & 36.27 \ & 22.00 \ & 14.89 \ & 62.42 \ & 22.80 \ & 25.77 \ & 69.35 \ & 33.02 \  & 0.08 s (GPU) \\
12 & GCSF~\cite{Cech2011} & 11.64 \ & 27.11 \ & 14.21 \ & 32.94 \ & 35.77 \ & 33.41 \ & 47.38 \ & 45.08 \ & 47.00 \ & 52.92 \ & 59.11 \ & 53.95 \  & 2.4 s \\
\end{tabular}
\end{minipage}
\vskip 1.5mm
\begin{minipage}{\linewidth}
\centering
\footnotesize
\includegraphics[width=0.16\linewidth]{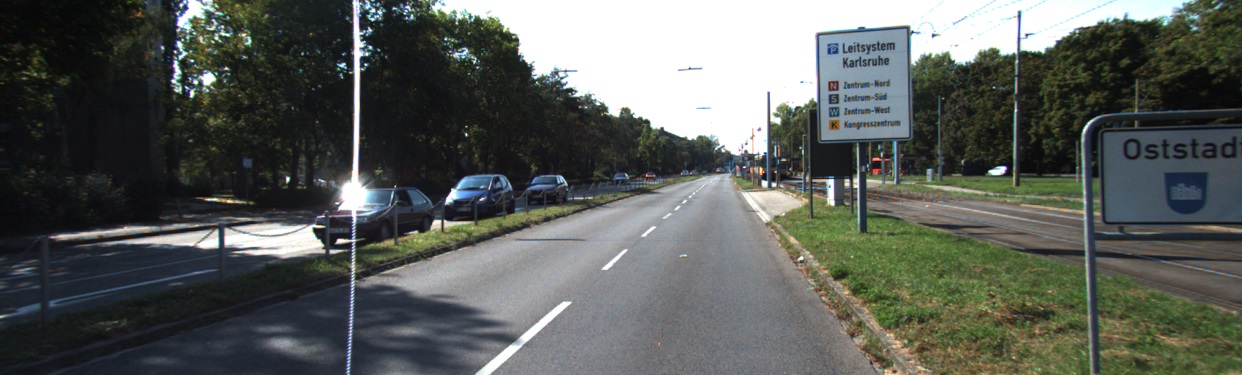}\hfill
\includegraphics[width=0.16\linewidth]{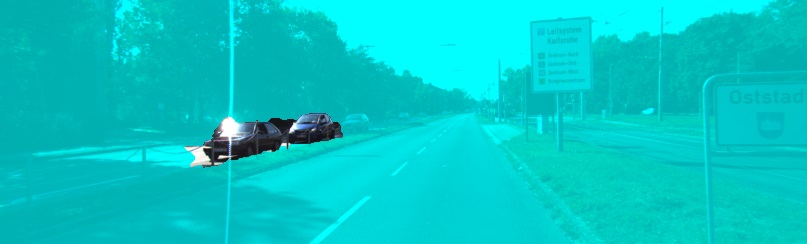}\hfill
\includegraphics[width=0.16\linewidth]{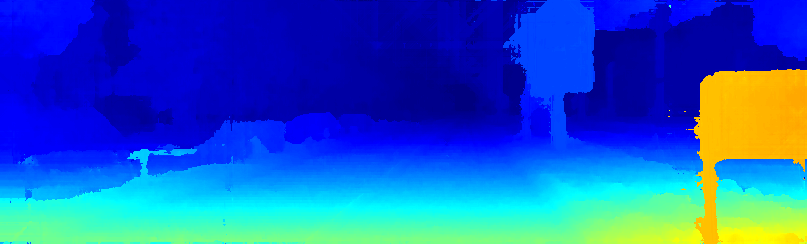}\hfill
\includegraphics[width=0.16\linewidth]{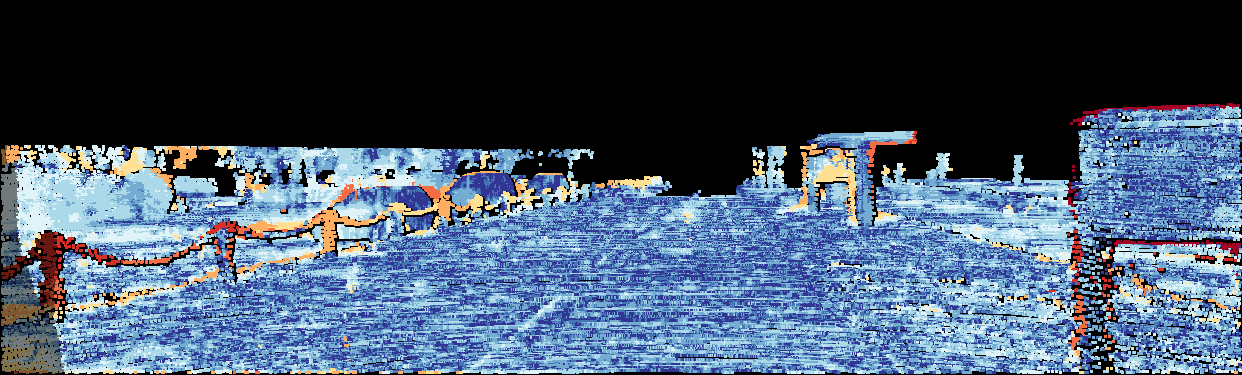}\hfill
\includegraphics[width=0.16\linewidth]{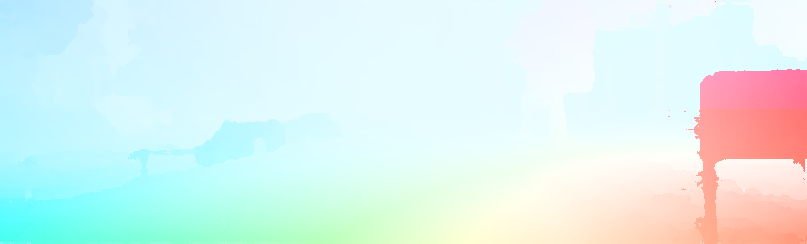}\hfill
\includegraphics[width=0.16\linewidth]{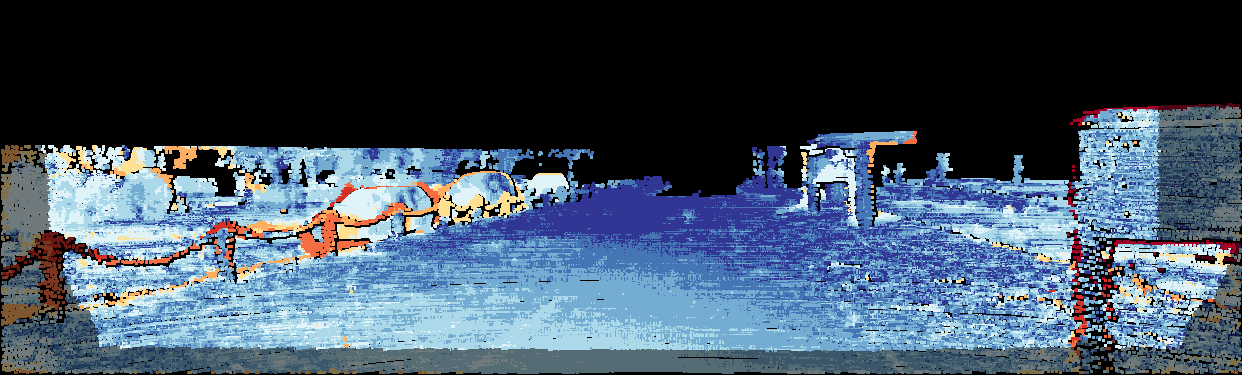}\\
\vskip 0.5mm
\includegraphics[width=0.16\linewidth]{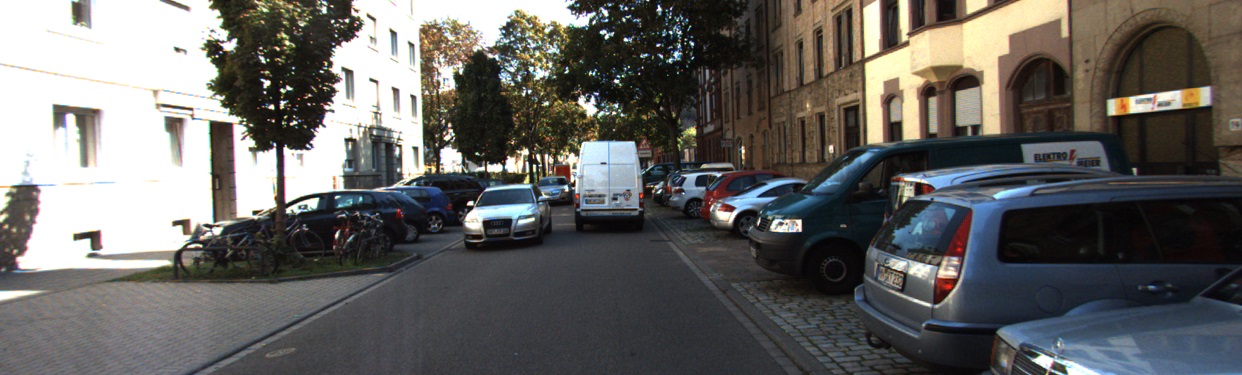}\hfill
\includegraphics[width=0.16\linewidth]{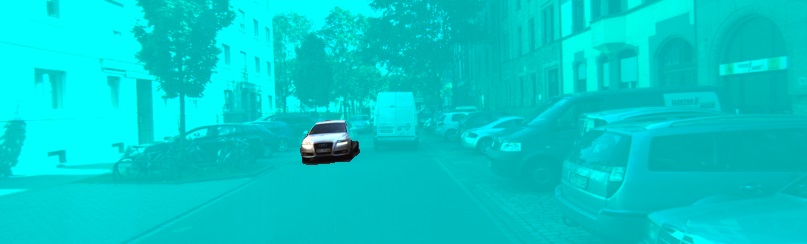}\hfill
\includegraphics[width=0.16\linewidth]{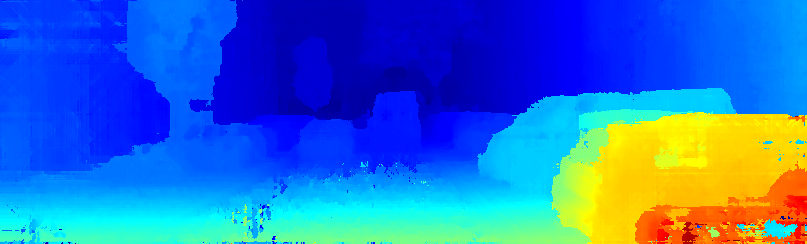}\hfill
\includegraphics[width=0.16\linewidth]{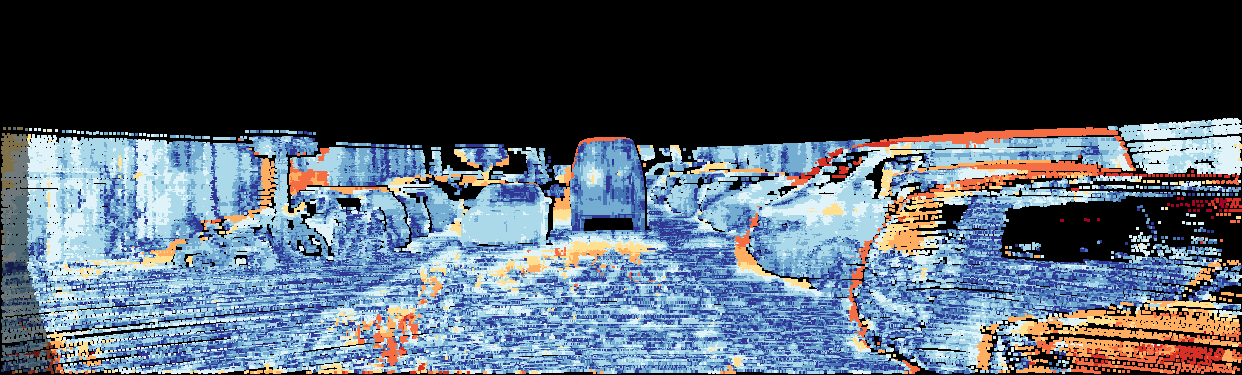}\hfill
\includegraphics[width=0.16\linewidth]{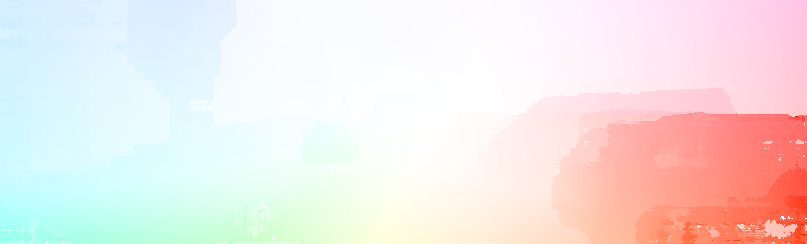}\hfill
\includegraphics[width=0.16\linewidth]{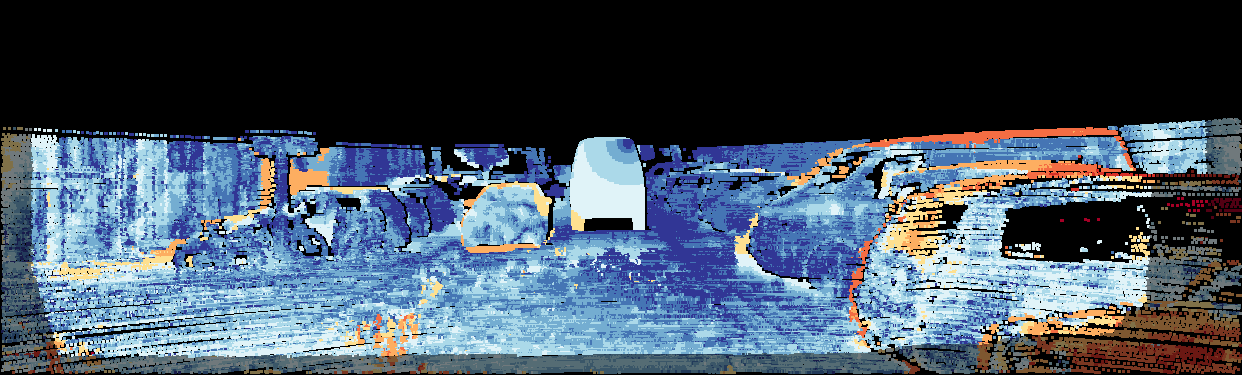}\\
\makebox[0.16\linewidth]{(a) Reference image}\hfill
\makebox[0.16\linewidth]{(b) Motion segmentation $\myS$}\hfill
\makebox[0.16\linewidth]{(c) Disparity map $\myD$}\hfill
\makebox[0.16\linewidth]{(d) Disparity error map}\hfill
\makebox[0.16\linewidth]{(e) Flow map $\myF$}\hfill
\makebox[0.16\linewidth]{(f) Flow error map}\hfill
\figcaption{Our results on KITTI testing sequences \textit{002} and \textit{006}. Black pixels in error heat maps indicate missing ground truth.}
\label{fig:kitti_results}
\end{minipage}
\vskip 3.5mm
\begin{minipage}{\linewidth}
\begin{minipage}{0.48\linewidth}
\begin{minipage}{\linewidth}
\caption{Disparity improvements by epipolar stereo.}
\label{tab:epipolar}
\scriptsize
\vskip 0.5mm
\begin{tabular}{c|c|c|c|c|c|c}
 & \multicolumn{3}{c|}{all pixels} & \multicolumn{3}{c}{non-occluded pixels} \\ 
 & D1-bg & D1-fg & D1-all &  D1-bg & D1-fg & D1-all \\ \hline 
Binocular  ($\tilde{\myD}$) & 7.96 & 12.61 & 8.68 & 7.09 & 10.57 & 7.61 \\
Epipolar  (${\myD}$) & \textbf{5.82} & \textbf{10.34} & \textbf{6.51} & \textbf{5.57} & \textbf{8.84} & \textbf{6.06} \\ 
\end{tabular}
\end{minipage}
\vskip 4mm
\begin{minipage}{\linewidth}
\centering
\includegraphics[width=0.85\linewidth]{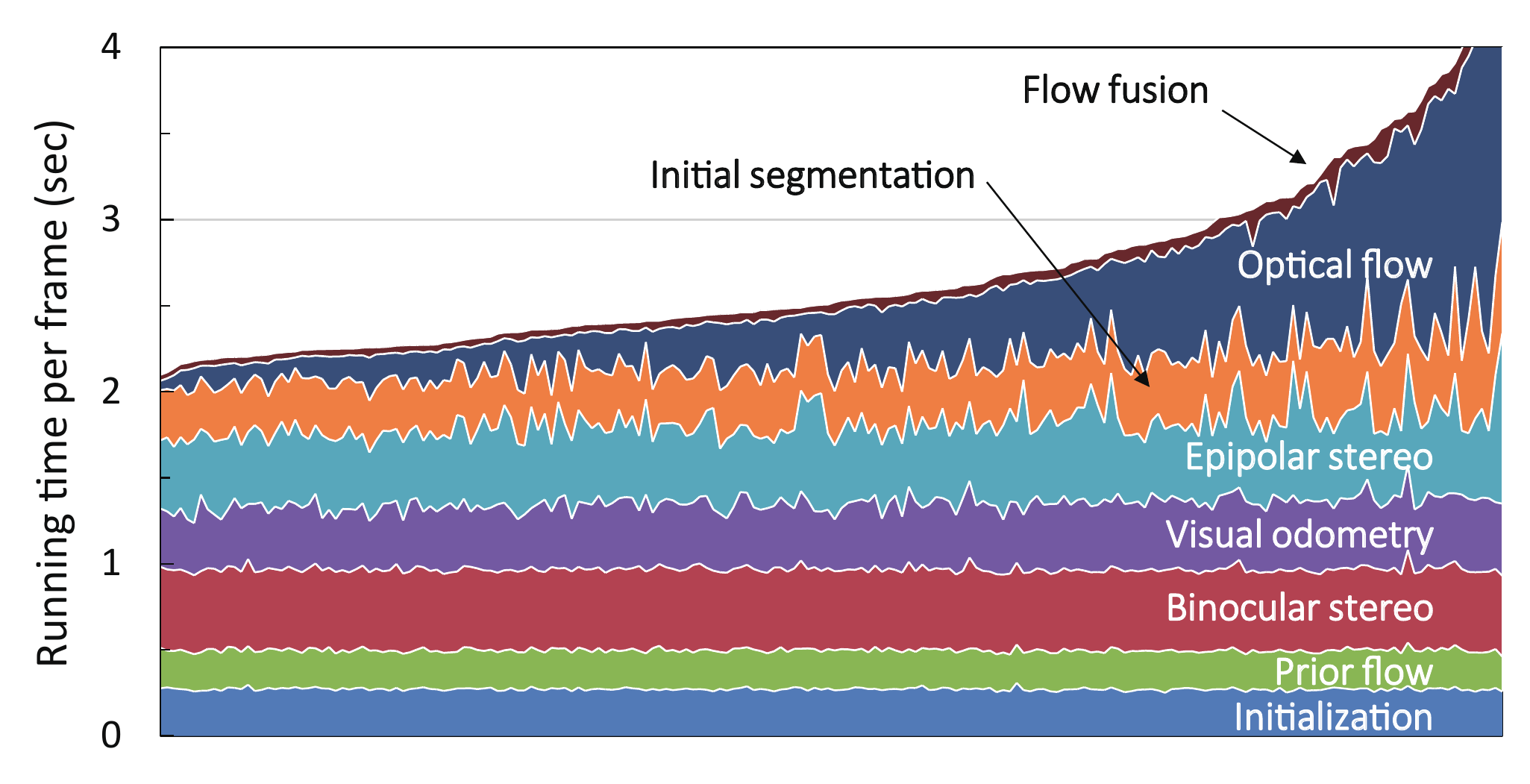}
\vskip -0.5mm
\figcaption{Running times on 200 sequences from KITTI. The average running time per-frame was 2.7 sec. Initialization includes edge extraction~\cite{Dollar2015}, superpixelization~\cite{VandenBergh2012} and feature tracking.}
\label{fig:running_time}
\end{minipage}
\end{minipage}
\hfill
\begin{minipage}{0.52\linewidth}
\centering
\caption{Sintel evaluation~\cite{Butler2012}: 
We show error rates ($\%$) for disparity (D1), flow (Fl), scene flow (SF) and motion segmentation (MS) averaged over the frames. 
Cell colors in OSF~\cite{Menze2015} and PRSM~\cite{Vogel2015} columns show performances relative to ours; blue shows where our method is better, red shows where it is worse.
We outperform OSF most of the time. 
}
\label{tab:sintel_scores}
\vskip 0.5mm
\includegraphics[width=0.90\linewidth]{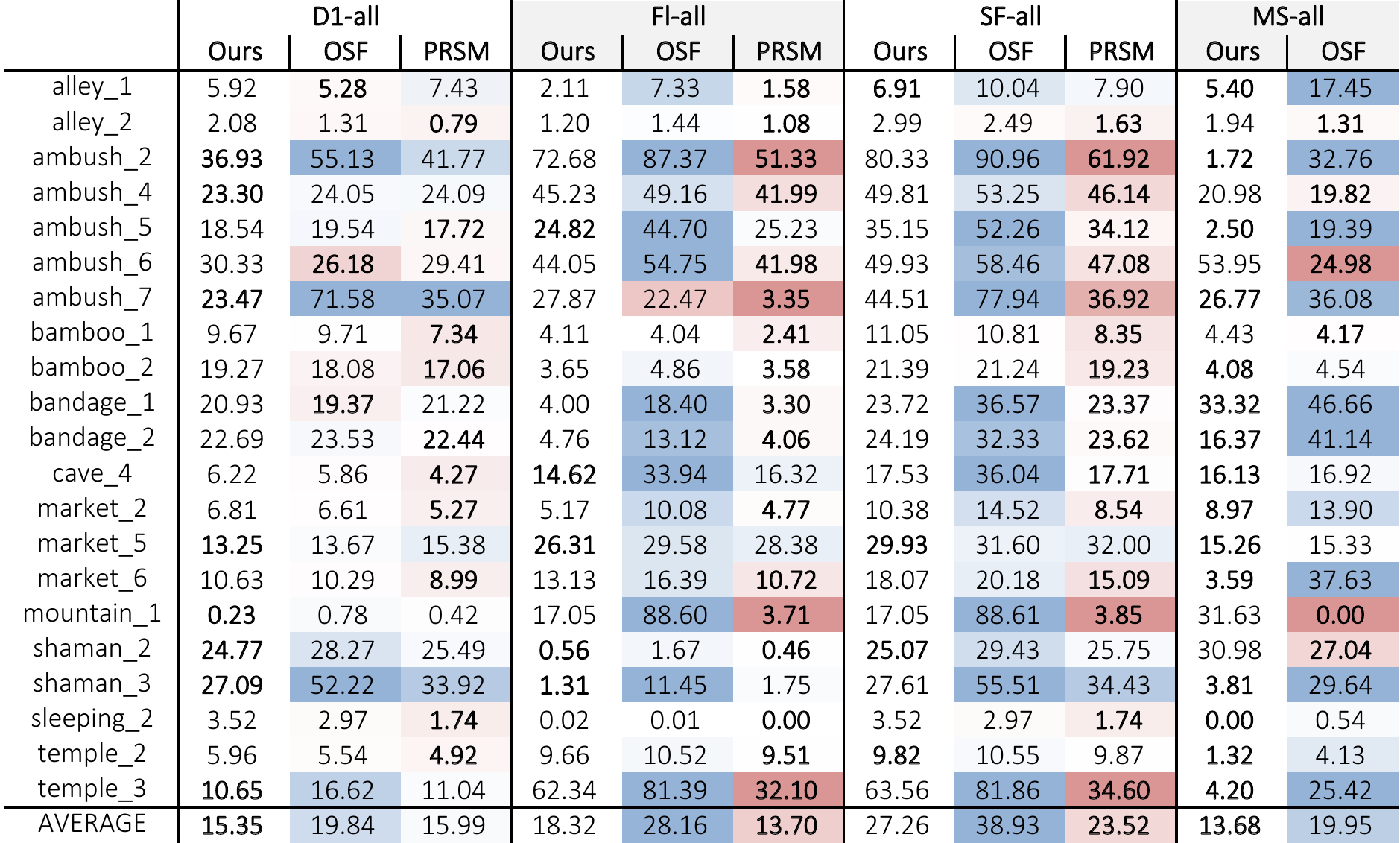}
\end{minipage}
\end{minipage}
\vskip 3.5mm
\begin{minipage}{\linewidth}
\scriptsize
\centering
\begin{overpic}[width=0.16\linewidth]{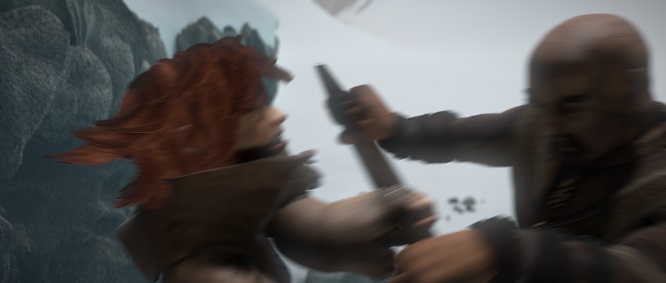}
\put(55,3){\textcolor{white}{\sf\textbf{ambush\_5}}}
\end{overpic}\hfill
\begin{overpic}[width=0.16\linewidth]{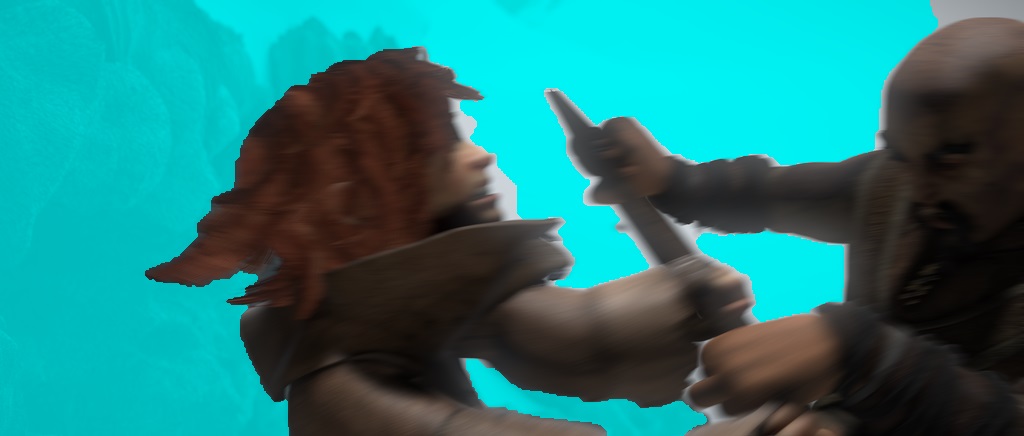}
\put(76,3){\textcolor{white}{\sf\textbf{Ours}}}
\end{overpic}\hfill
\begin{overpic}[width=0.16\linewidth]{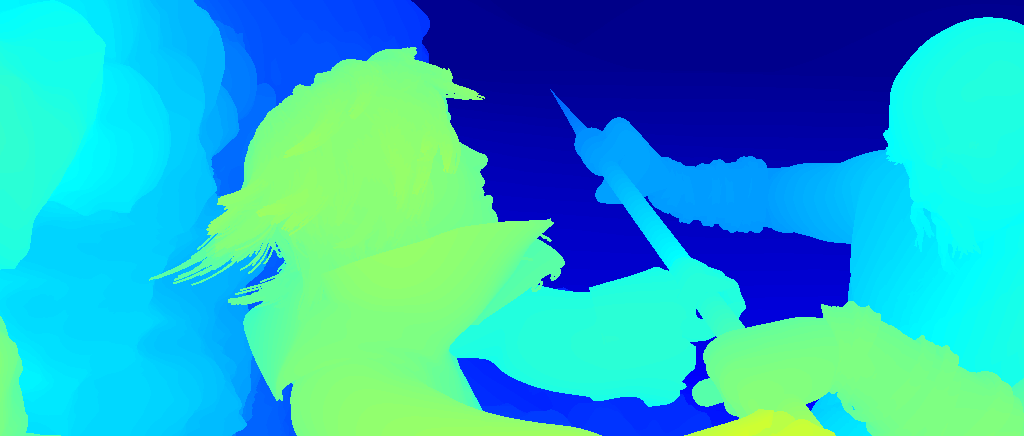}
\put(84,3){\textcolor{black}{\sf\textbf{GT}}}
\end{overpic}\hfill
\begin{overpic}[width=0.16\linewidth]{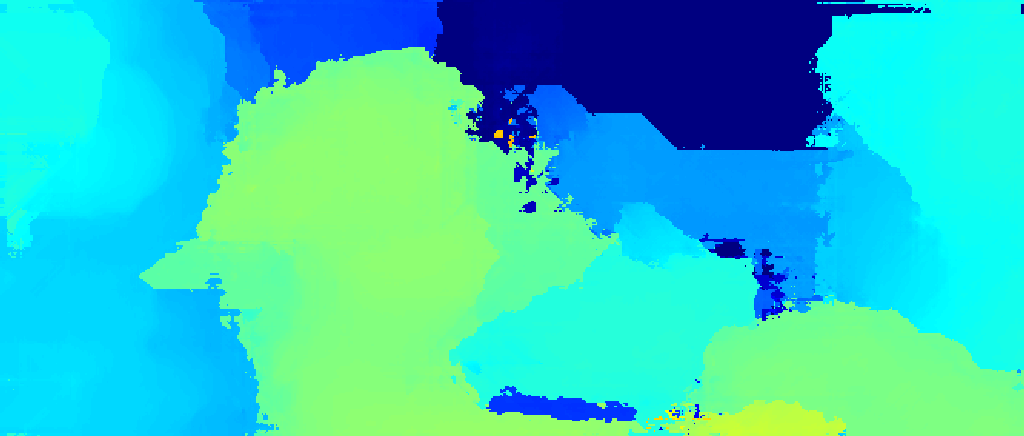}
\put(76,3){\textcolor{black}{\sf\textbf{Ours}}}
\end{overpic}\hfill
\begin{overpic}[width=0.16\linewidth]{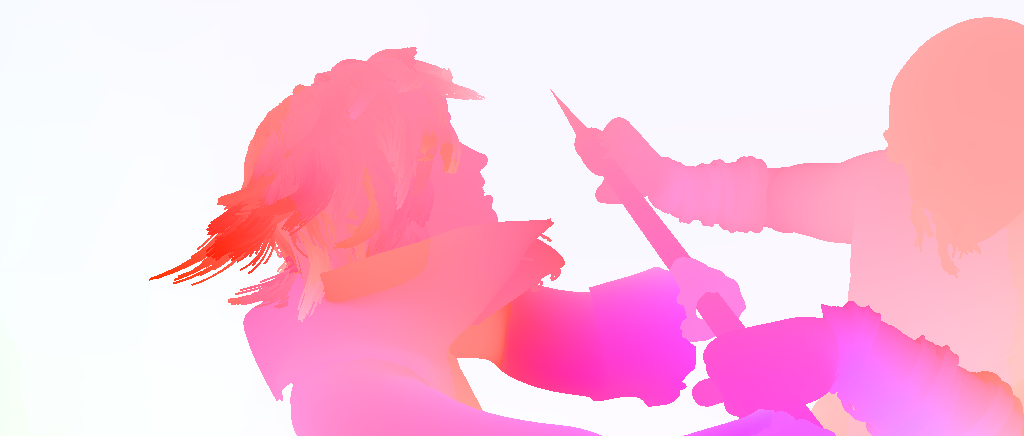}
\put(84,3){\textcolor{white}{\sf\textbf{GT}}}
\end{overpic}\hfill
\begin{overpic}[width=0.16\linewidth]{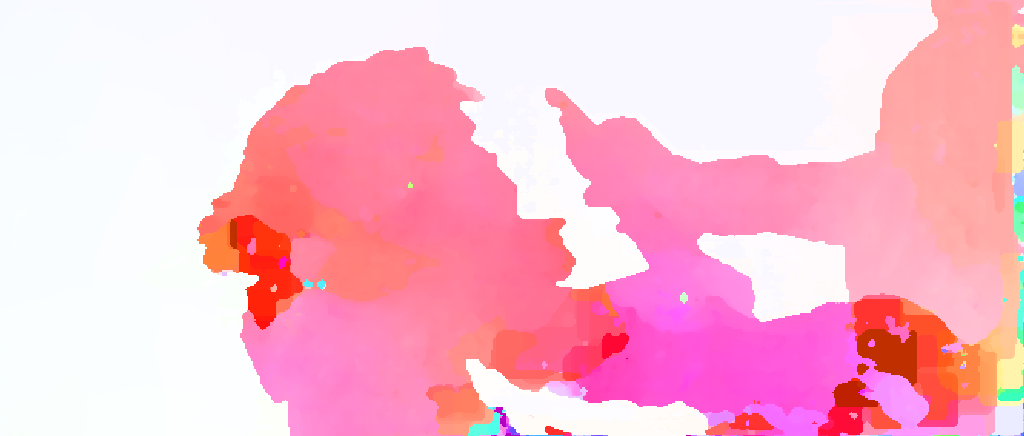}
\put(76,3){\textcolor{white}{\sf\textbf{Ours}}}
\end{overpic}
\vskip 0.5mm
\begin{overpic}[width=0.16\linewidth]{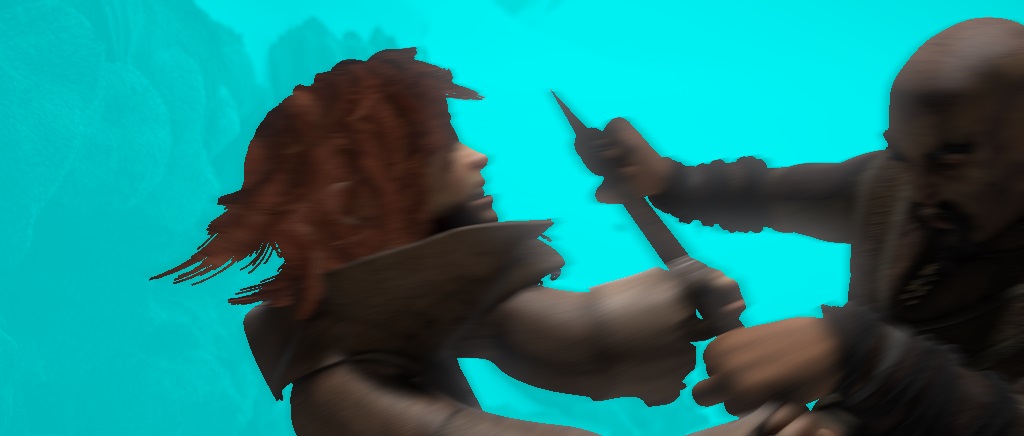}
\put(84,3){\textcolor{white}{\sf\textbf{GT}}}
\end{overpic}\hfill
\begin{overpic}[width=0.16\linewidth]{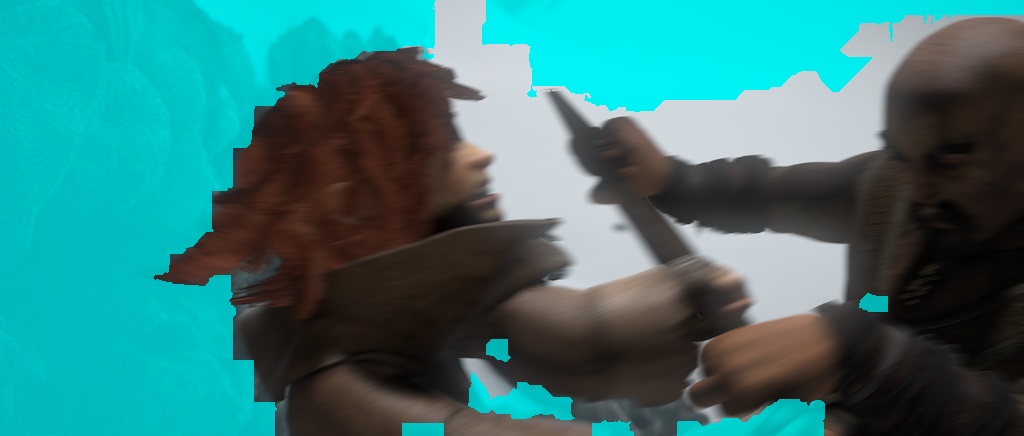}
\put(77,3){\textcolor{white}{\sf\textbf{OSF}}}
\end{overpic}\hfill
\begin{overpic}[width=0.16\linewidth]{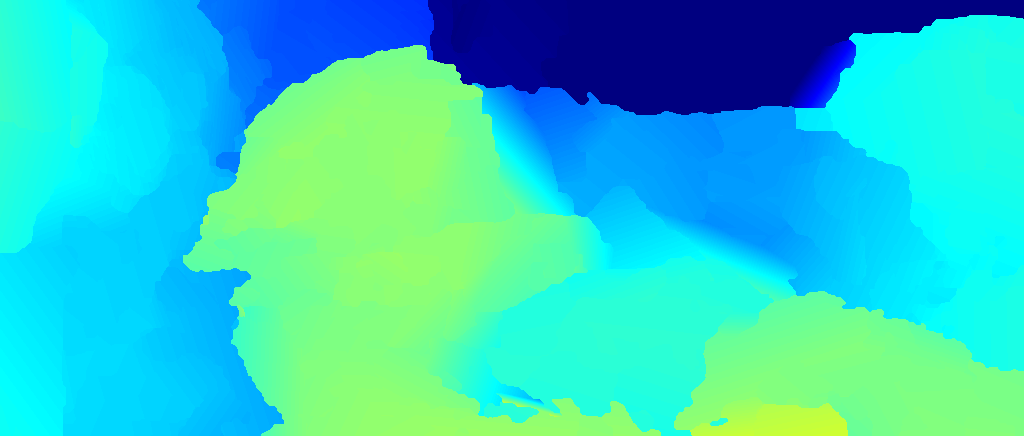}
\put(71,3){\textcolor{black}{\sf\textbf{PRSM}}}
\end{overpic}\hfill
\begin{overpic}[width=0.16\linewidth]{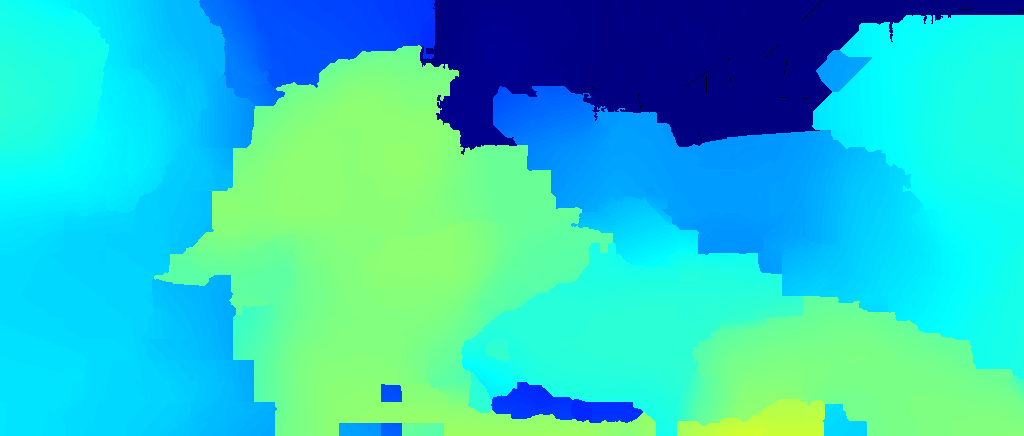}
\put(77,3){\textcolor{black}{\sf\textbf{OSF}}}
\end{overpic}\hfill
\begin{overpic}[width=0.16\linewidth]{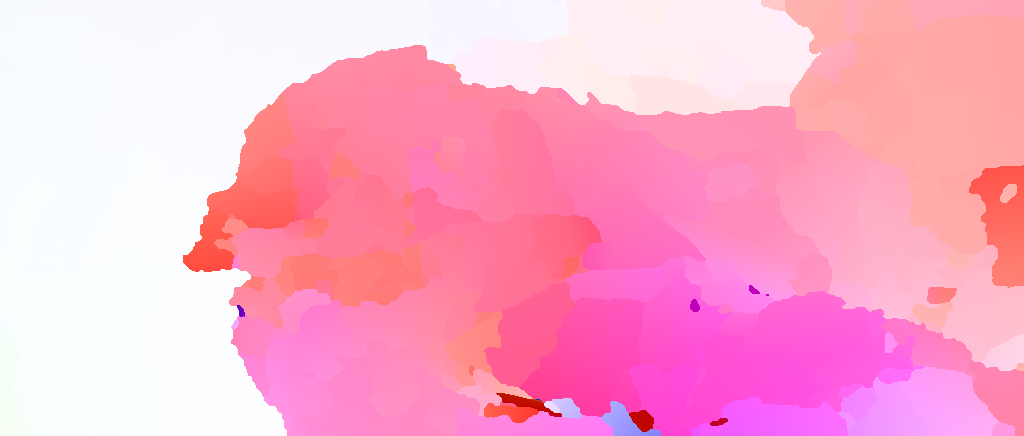}
\put(71,3){\textcolor{white}{\sf\textbf{PRSM}}}
\end{overpic}\hfill
\begin{overpic}[width=0.16\linewidth]{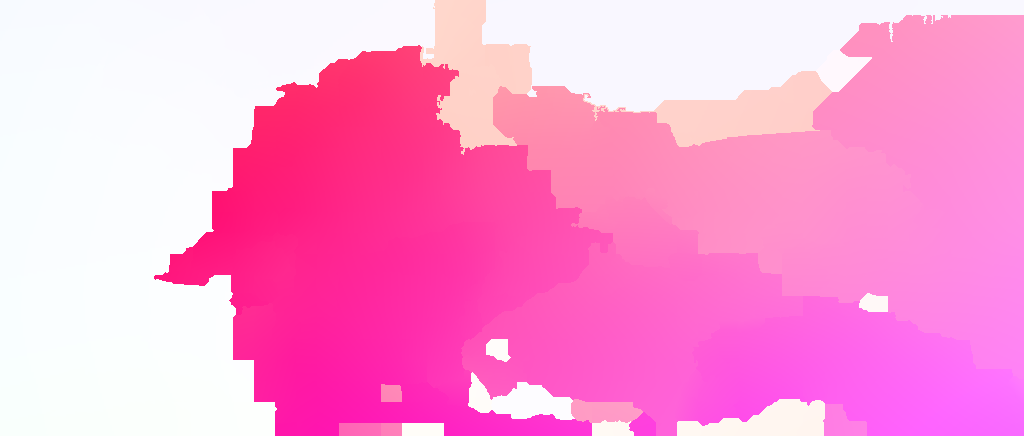}
\put(77,3){\textcolor{white}{\sf\textbf{OSF}}}
\end{overpic}
\vskip 1.0mm
\begin{overpic}[width=0.16\linewidth]{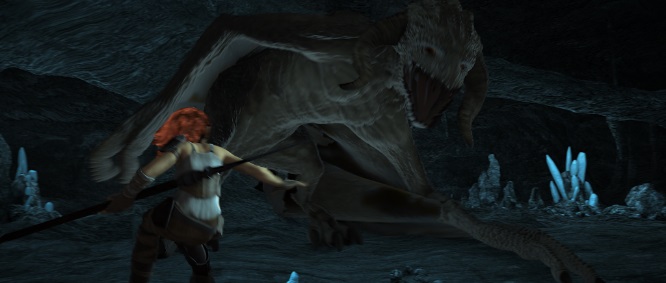}
\put(68,3){\textcolor{white}{\sf\textbf{cave\_4}}}
\end{overpic}\hfill
\begin{overpic}[width=0.16\linewidth]{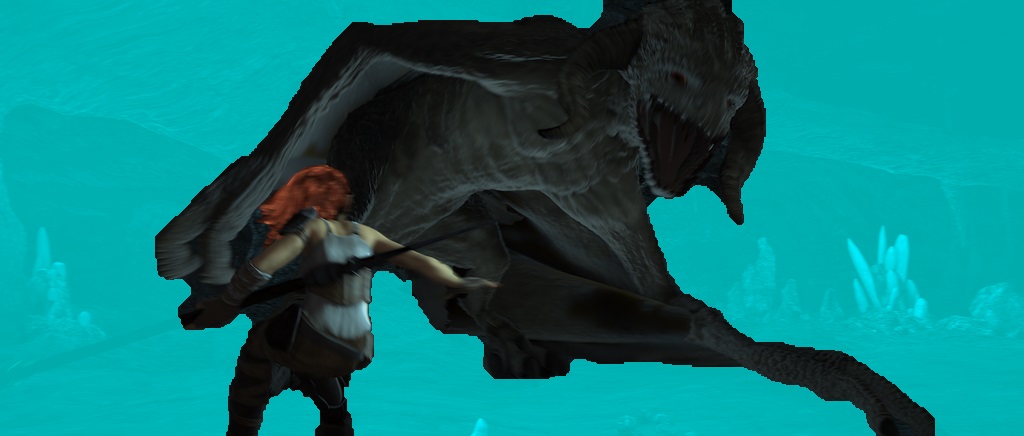}
\put(76,3){\textcolor{white}{\sf\textbf{Ours}}}
\end{overpic}\hfill
\begin{overpic}[width=0.16\linewidth]{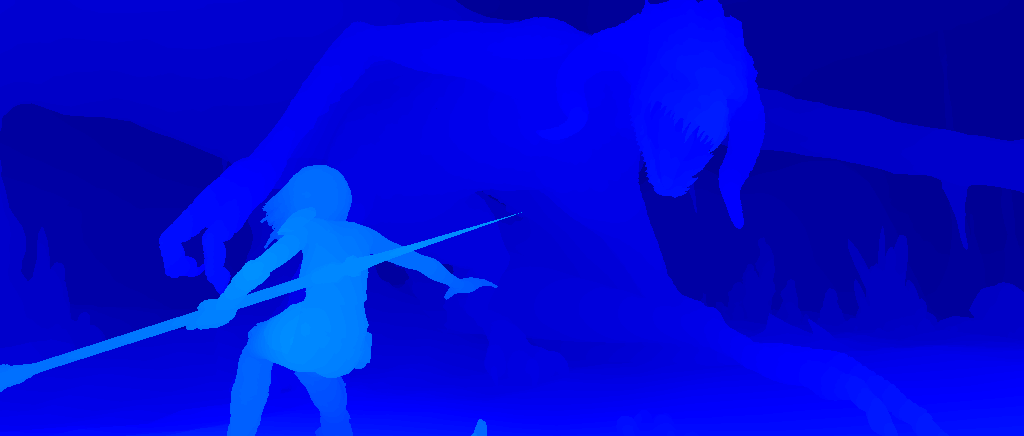}
\put(84,3){\textcolor{white}{\sf\textbf{GT}}}
\end{overpic}\hfill
\begin{overpic}[width=0.16\linewidth]{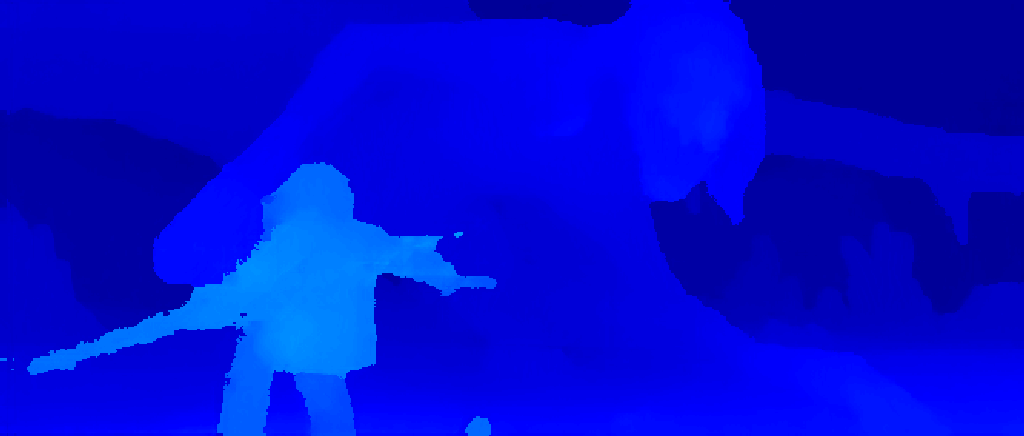}
\put(76,3){\textcolor{white}{\sf\textbf{Ours}}}
\end{overpic}\hfill
\begin{overpic}[width=0.16\linewidth]{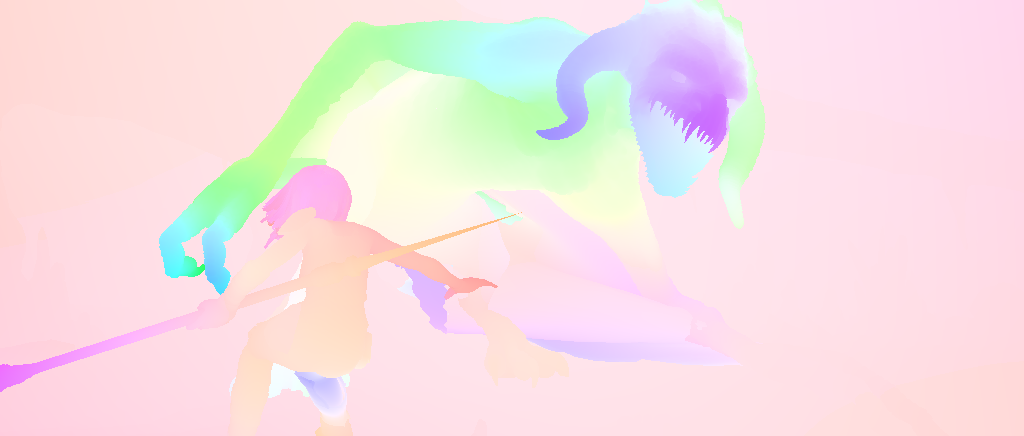}
\put(84,3){\textcolor{black}{\sf\textbf{GT}}}
\end{overpic}\hfill
\begin{overpic}[width=0.16\linewidth]{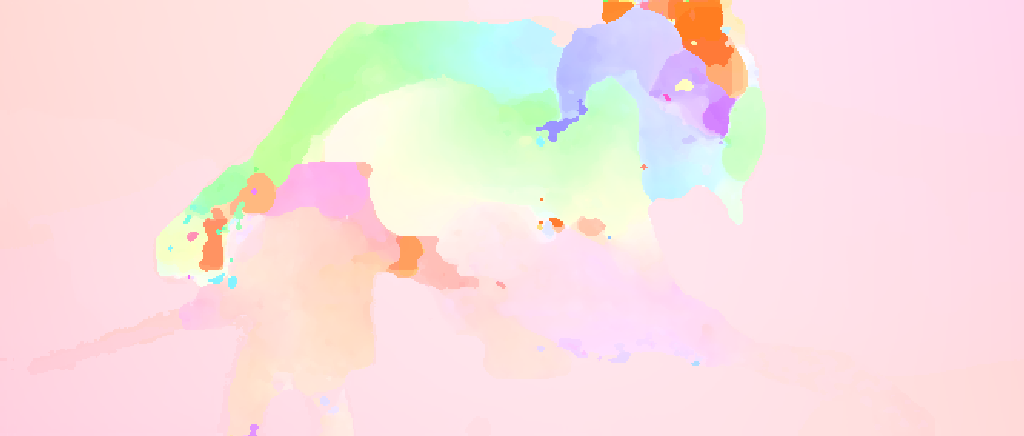}
\put(76,3){\textcolor{black}{\sf\textbf{Ours}}}
\end{overpic}
\vskip 0.5mm
\begin{overpic}[width=0.16\linewidth]{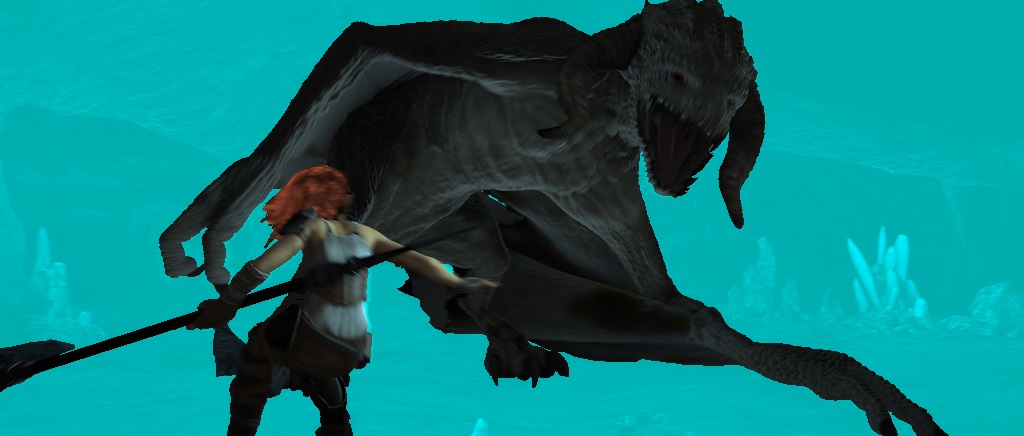}
\put(84,3){\textcolor{white}{\sf\textbf{GT}}}
\end{overpic}\hfill
\begin{overpic}[width=0.16\linewidth]{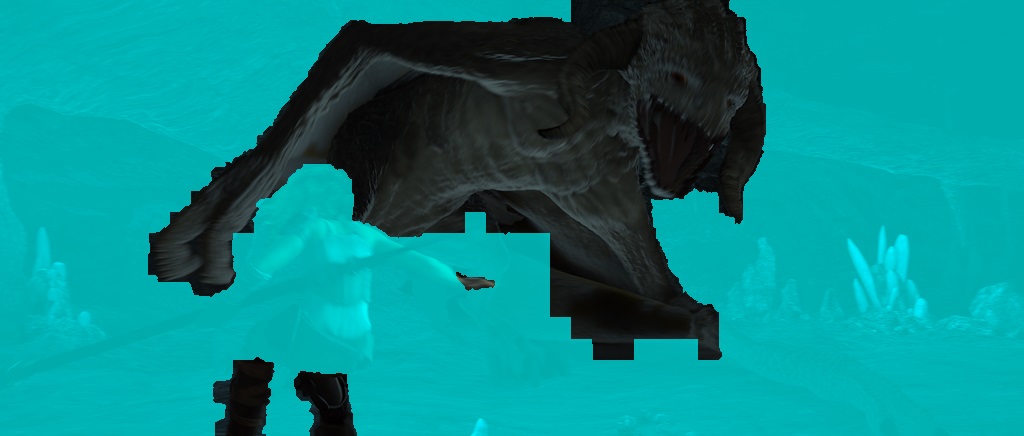}
\put(77,3){\textcolor{white}{\sf\textbf{OSF}}}
\end{overpic}\hfill
\begin{overpic}[width=0.16\linewidth]{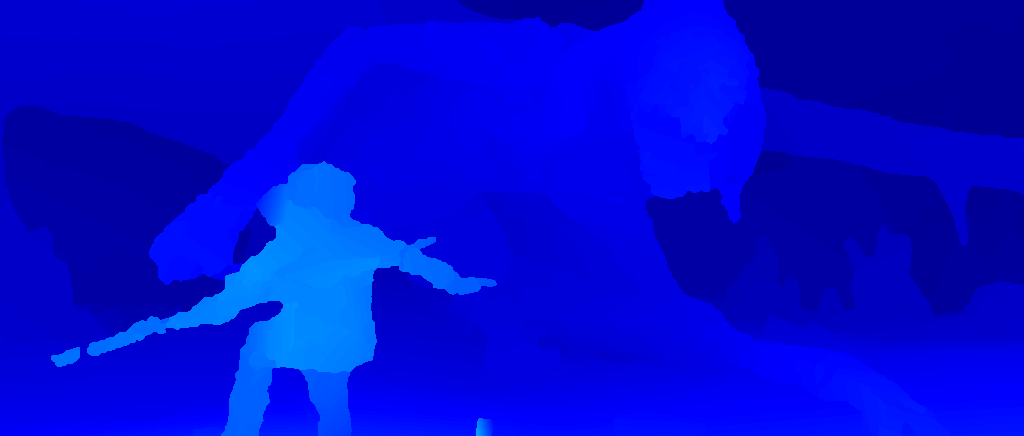}
\put(71,3){\textcolor{white}{\sf\textbf{PRSM}}}
\end{overpic}\hfill
\begin{overpic}[width=0.16\linewidth]{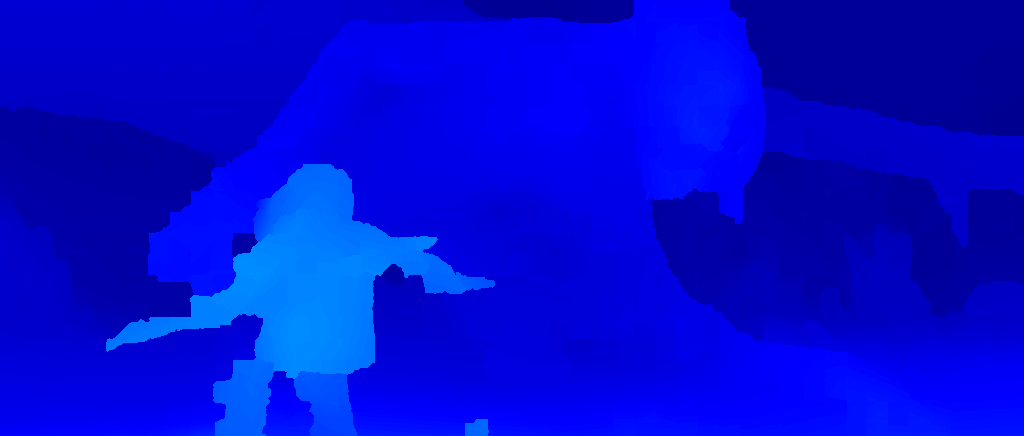}
\put(77,3){\textcolor{white}{\sf\textbf{OSF}}}
\end{overpic}\hfill
\begin{overpic}[width=0.16\linewidth]{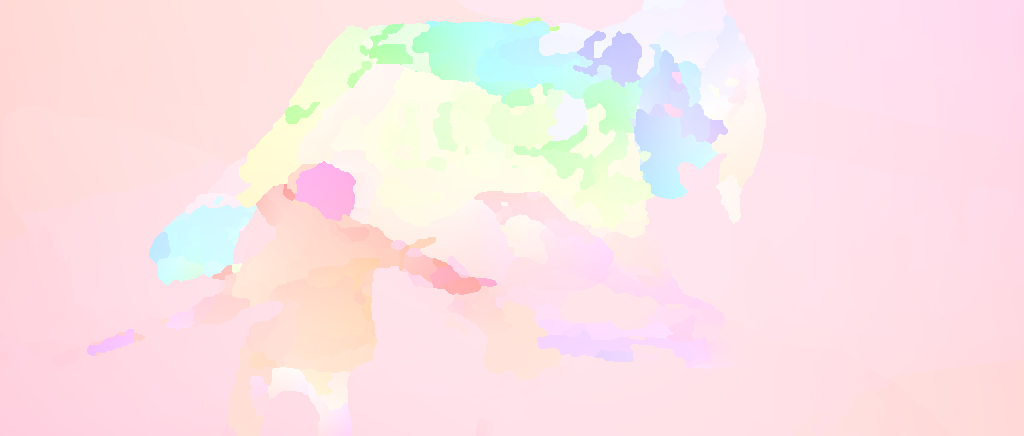}
\put(71,3){\textcolor{black}{\sf\textbf{PRSM}}}
\end{overpic}\hfill
\begin{overpic}[width=0.16\linewidth]{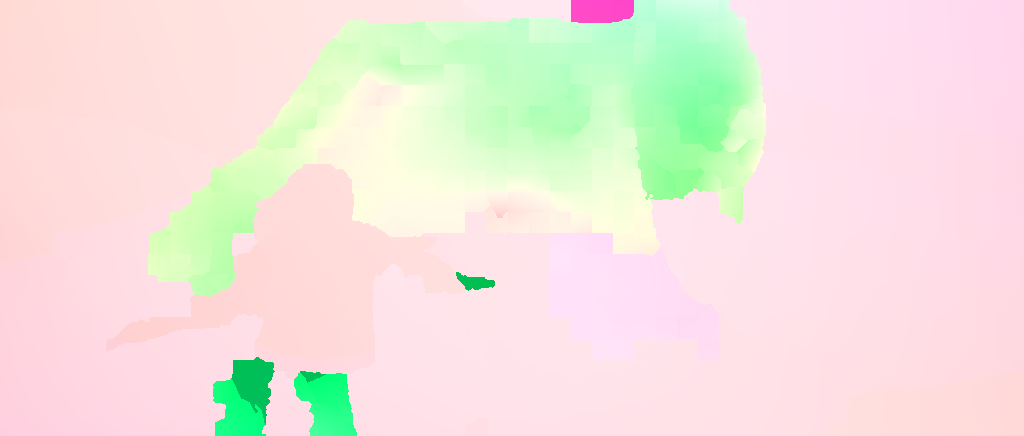}
\put(77,3){\textcolor{black}{\sf\textbf{OSF}}}
\end{overpic}
\vskip 1.0mm
\begin{overpic}[width=0.16\linewidth]{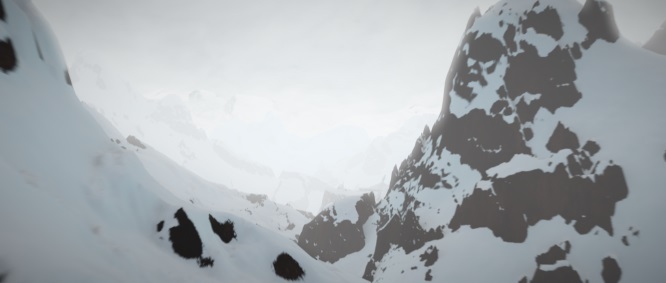}
\put(48,3){\textcolor{white}{\sf\textbf{mountain\_1}}}
\end{overpic}\hfill
\begin{overpic}[width=0.16\linewidth]{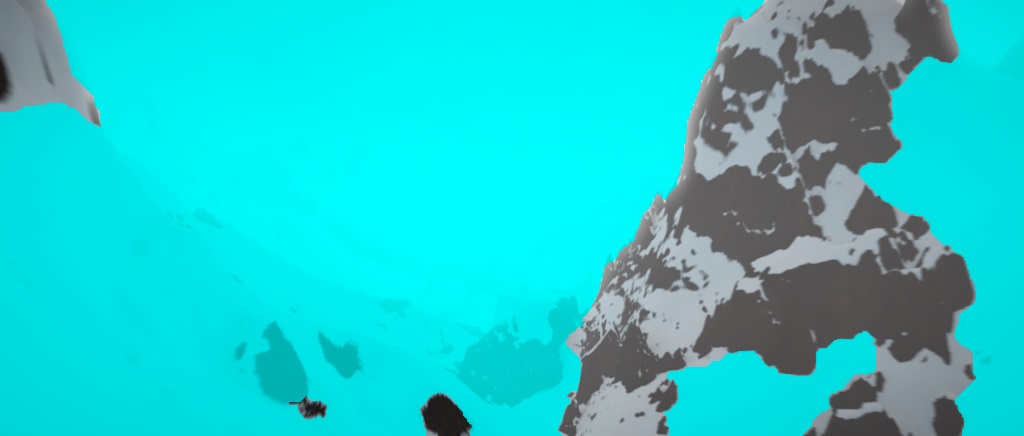}
\put(76,3){\textcolor{white}{\sf\textbf{Ours}}}
\end{overpic}\hfill
\begin{overpic}[width=0.16\linewidth]{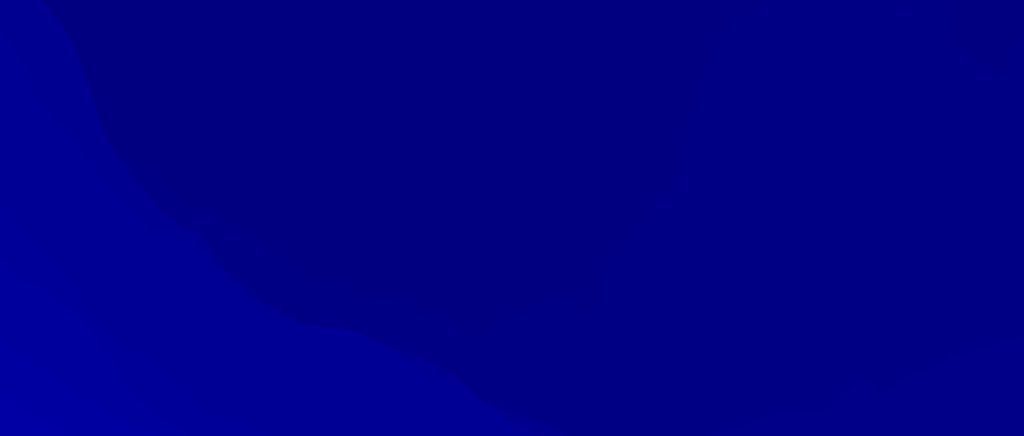}
\put(84,3){\textcolor{white}{\sf\textbf{GT}}}
\end{overpic}\hfill
\begin{overpic}[width=0.16\linewidth]{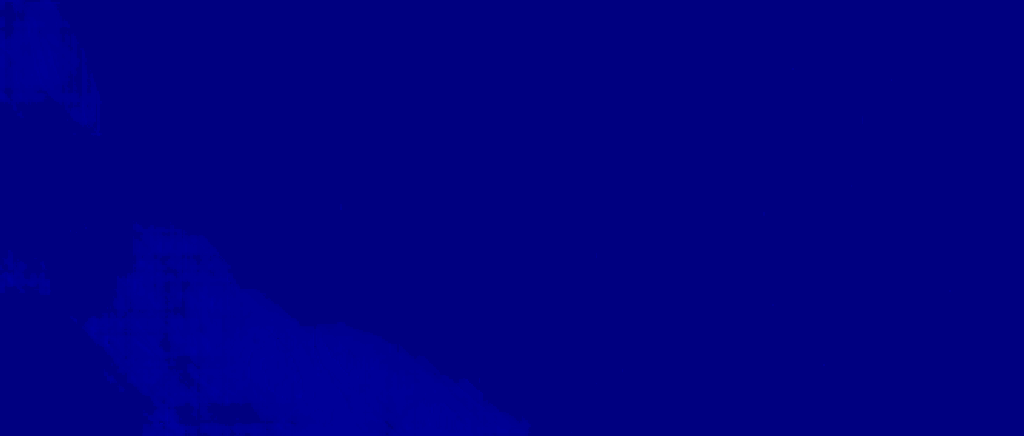}
\put(76,3){\textcolor{white}{\sf\textbf{Ours}}}
\end{overpic}\hfill
\begin{overpic}[width=0.16\linewidth]{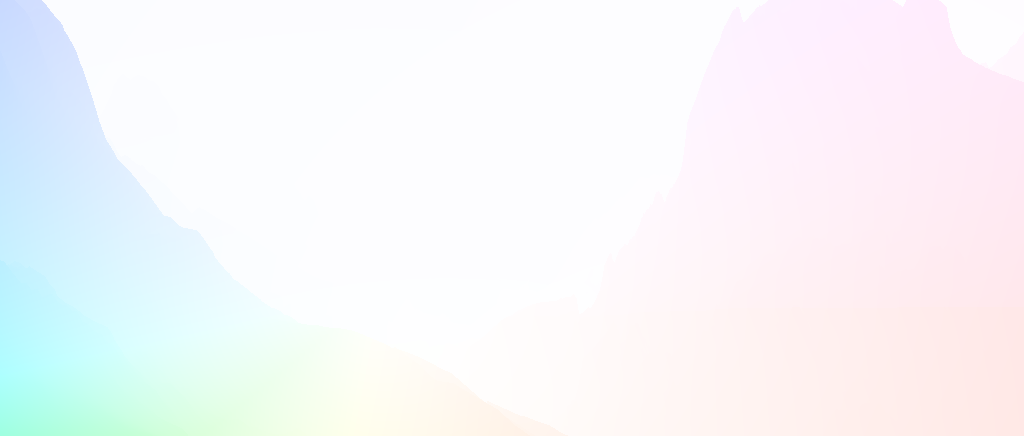}
\put(84,3){\textcolor{black}{\sf\textbf{GT}}}
\end{overpic}\hfill
\begin{overpic}[width=0.16\linewidth]{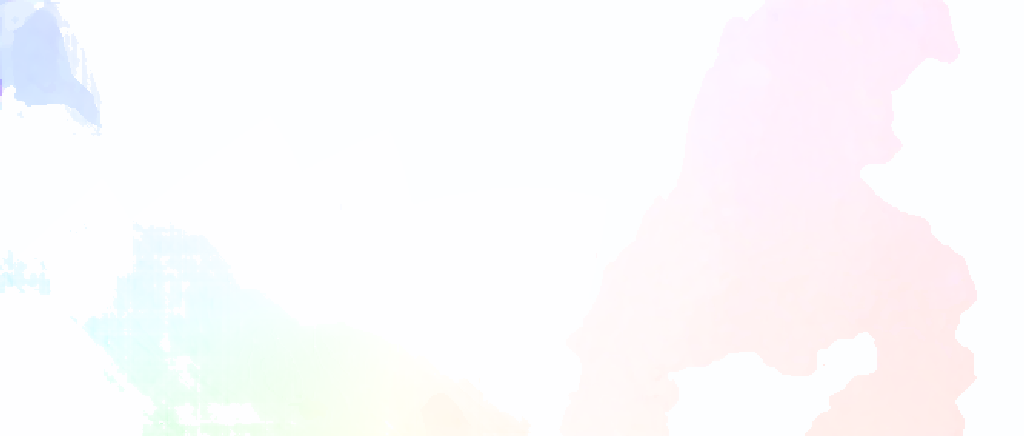}
\put(76,3){\textcolor{black}{\sf\textbf{Ours}}}
\end{overpic}
\vskip 0.5mm
\begin{overpic}[width=0.16\linewidth]{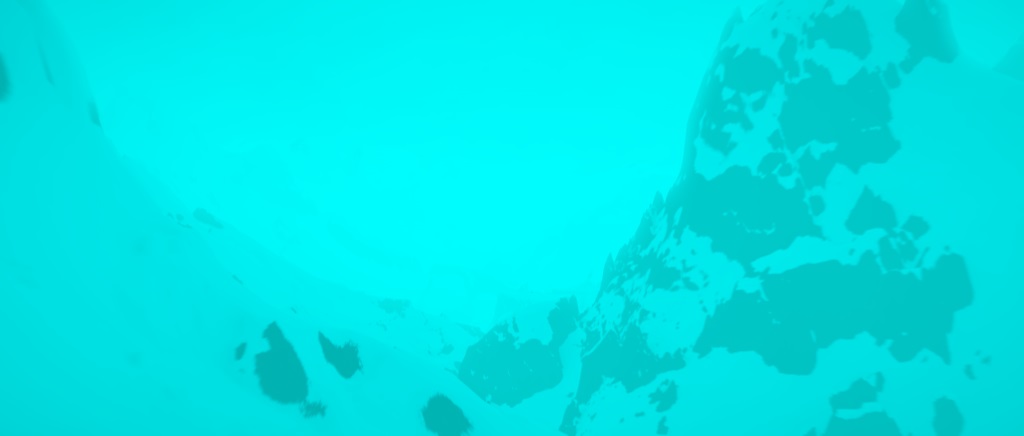}
\put(84,3){\textcolor{white}{\sf\textbf{GT}}}
\end{overpic}\hfill
\begin{overpic}[width=0.16\linewidth]{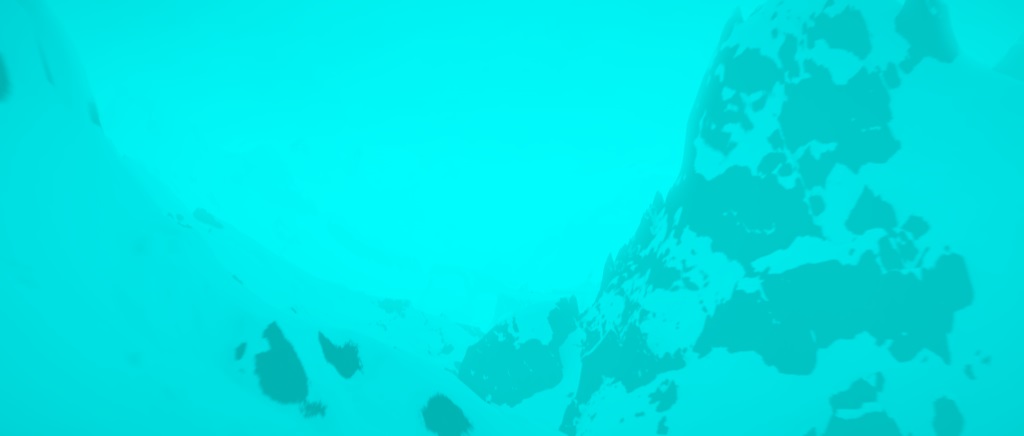}
\put(77,3){\textcolor{white}{\sf\textbf{OSF}}}
\end{overpic}\hfill
\begin{overpic}[width=0.16\linewidth]{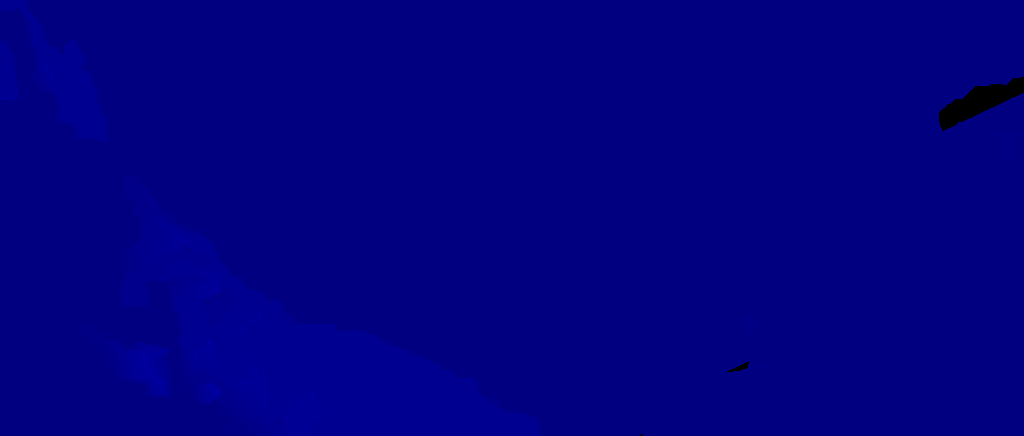}
\put(71,3){\textcolor{white}{\sf\textbf{PRSM}}}
\end{overpic}\hfill
\begin{overpic}[width=0.16\linewidth]{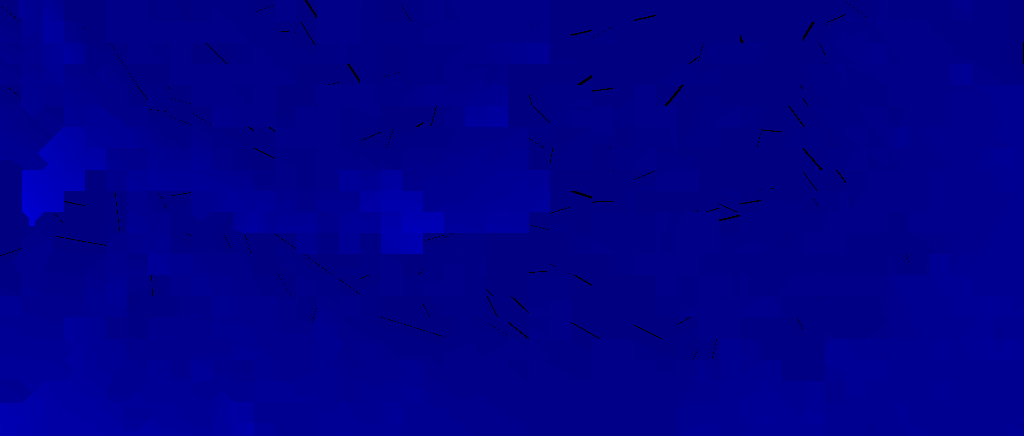}
\put(77,3){\textcolor{white}{\sf\textbf{OSF}}}
\end{overpic}\hfill
\begin{overpic}[width=0.16\linewidth]{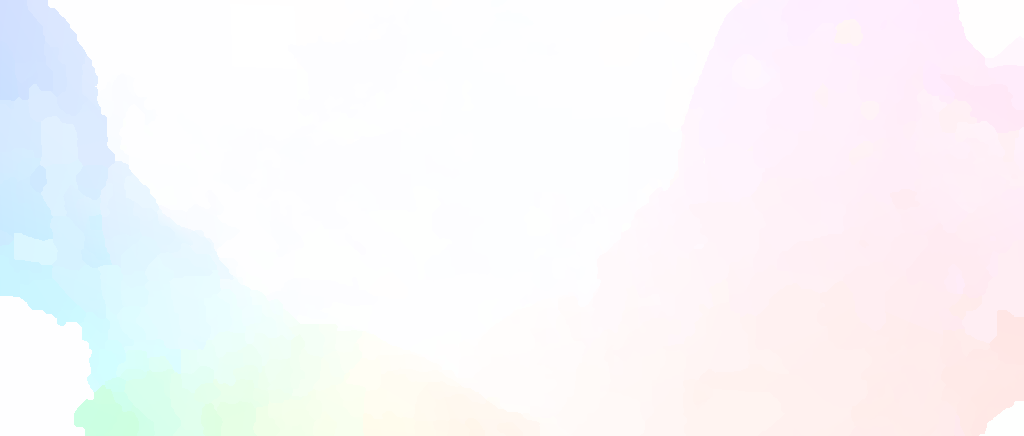}
\put(71,3){\textcolor{black}{\sf\textbf{PRSM}}}
\end{overpic}\hfill
\begin{overpic}[width=0.16\linewidth]{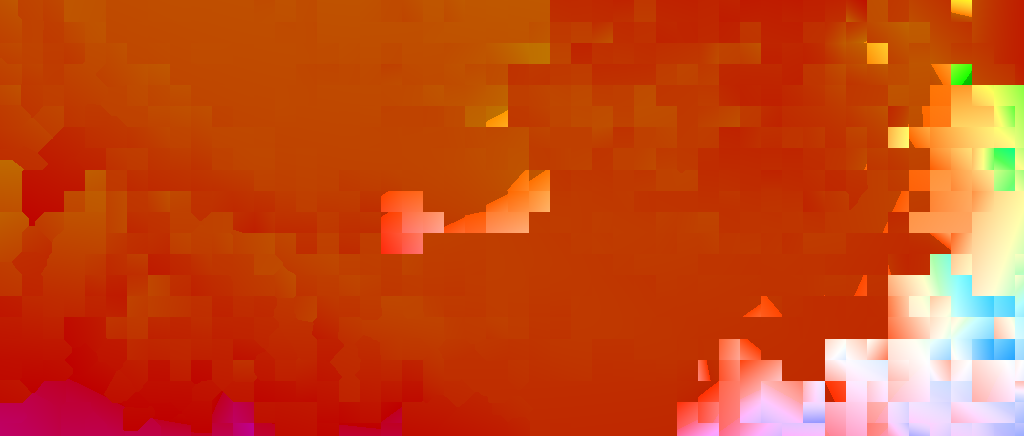}
\put(77,3){\textcolor{black}{\sf\textbf{OSF}}}
\end{overpic}
\footnotesize
\makebox[0.32\textwidth]{Reference images / motion segmentation}\hfill
\makebox[0.32\textwidth]{Disparity maps}\hfill
\makebox[0.32\textwidth]{Flow maps}\hfill
\figcaption{Comparisons on \textsl{ambush\_5}, \textsl{cave\_4} and \textsl{mountain\_1} from Sintel: [\textsc{Left}] Motion segmentation results -- ours, OSF and ground truth. [\textsc{Middle}] Disparity and [\textsc{Right}] Flow maps estimated by our method, PRSM and OSF and the ground truth versions.}
\label{fig:sintel_results}
\end{minipage}
\end{table*}

\section{Conclusions}
We proposed an efficient scene flow method that unifies dense stereo, optical flow, visual odometry, and motion segmentation estimation. Even though simple optimization methods were used in our technique, the unified framework led to higher overall accuracy and efficiency.
Our method is currently ranked third on the KITTI 2015 scene flow benchmark after PRSM~\cite{Vogel2015} and OSF~\cite{Menze2015} but is 1--3 orders of magnitude faster than the top six methods. On challenging Sintel sequences, our method outperforms OSF~\cite{Menze2015} and is close to PRSM~\cite{Vogel2015} in terms of accuracy. Our efficient method could be used to initialize PRSM~\cite{Vogel2015} to improve its convergence speed. We hope it will enable new, practical applications of scene flow.

{\small
\bibliographystyle{ieee}
\bibliography{references}
}

\newpage
\clearpage
\onecolumn
\setcounter{page}{1}
\setcounter{section}{0}
\setcounter{figure}{0}
\setcounter{table}{0}
\setcounter{equation}{0}
\setcounter{footnote}{0}
\renewcommand{\thepage}{A\arabic{page}}
\renewcommand{\thesection}{\Alph{section}}
\renewcommand{\thefigure}{A\arabic{figure}}
\renewcommand{\thetable}{A\arabic{table}}
\renewcommand{\theequation}{A\arabic{equation}}
\setstretch{1.2}

\makeatletter
\def\@thanks{}
\makeatother

\title{{Fast Multi-frame Stereo Scene Flow with Motion Segmentation\\--- Supplementary  Material ---}}

\author{Tatsunori Taniai\\ 
RIKEN AIP\\
\and
Sudipta N. Sinha\\
Microsoft Research\\
\and
Yoichi Sato\\
The University of Tokyo
}

\maketitle


\pagenumbering{gobble}
In the supplementary material we present details of our SGM stereo and flow implementations (used in Sec.~\ref{sec:stereo} and Sec.~\ref{sec:optflow}) as well as the segmentation ground prior (used in Sec.~\ref{sec:details}) that were omitted from the main paper due to the limit on page length. We also discuss parameter settings and their effects on our method.
Note that we review the SGM algorithm as proposed by Hirschmuller~\cite{Hirschmuller2008}, but describe the algorithm using our own notation to be consistent with the main paper.
We also provide additional qualitative results and comparisons with state-of-the-art methods in the supplementary video.

\section{SGM Stereo}
\label{sec:supp_sgm_stereo}
In the binocular and epipolar stereo stages (Sec.~\ref{sec:stereo} and \ref{sec:epistereo}), we solve stereo matching problems using the semi-global matching (SGM) algorithm~\cite{Hirschmuller2008}.
Here, stereo matching is cast as a discrete labeling problem, where we estimate the disparity map $\myD_\myp = \myD(\myp) : \Omega \to D$  (where $D = \{D_\text{min}, \cdots, D_\text{max}\}$ is the disparity range) that minimizes the following 2D Markov random field (MRF) based energy function.
\begin{equation}
E_{\text{stereo}}(\myD) = \sum_{\myp \in \Omega} C_\myp(\myD_\myp) +  \sum_{(\mathbf{p,q}) \in N} c V_\mathbf{pq}(\myD_\myp, \myD_\mathbf{q}). \label{eq:stereo}
\end{equation}
Here, $C_\myp(\myD_\myp)$ is the unary data term that evaluates photo-consistencies between the pixel $\myp$ in the left image $I^0$ and its corresponding pixel $\myp' = \myp - (\myD_\myp, 0)^T$ at the disparity $\myD_\myp$ in the right image $I^1$.
$V_\mathbf{pq}(\myD_\myp, \myD_\mathbf{q})$ is the pairwise smoothness term defined for neighboring pixel pairs $(\mathbf{p,q}) \in N$ on the 8-connected pixel grid. In SGM, this term is usually defined as
\begin{eqnarray}
V_\mathbf{pq}(\myD_\myp, \myD_\myq)=\left\{ \begin{array}{ll}
0 & \text{if} \;\; \myD_\myp = \myD_\myq \\
P_1 & \text{if} \;\; |\myD_\myp - \myD_\myq|=1 \\
P_2 & \text{otherwise} \\
\end{array} \right..
\end{eqnarray}
Here, $P_1$ and $P_2$ ($0 < P_1 < P_2$) are smoothness penalties.
The coefficient $c$ in Eq.~(\ref{eq:stereo}) is described later.

While the exact inference of Eq.~(\ref{eq:stereo}) is NP-hard,
SGM decomposes the 2D MRF into many 1D MRFs along 8 cardinal directions $\mathbf{r}$ and minimizes them using dynamic programming~\cite{Hirschmuller2008}. This is done by recursively updating the following cost arrays $L_\mathbf{r}(\mathbf{p}, d)$ along 1D scan lines in the directions $\mathbf{r}$ from the image boundary pixels.
\begin{eqnarray}
L_\mathbf{r}(\mathbf{p}, d) = C_\mathbf{p}(d) + \min_{d' \in D} \left[ L_\mathbf{r}(\mathbf{p} - \mathbf{r}, d') + V_\mathbf{pq}(d, d') \right] - \min_{d' \in D}  L_\mathbf{r}(\mathbf{p} - \mathbf{r}, d'). \label{eq:linecost}
\end{eqnarray}
Here, by introducing the following normalized scan-line costs
\begin{eqnarray}
\bar{L}_\mathbf{r}(\mathbf{p}, d) = L_\mathbf{r}(\mathbf{p}, d) - \min_{d' \in D}  L_\mathbf{r}(\mathbf{p}, d'), \label{eq:nlinecost}
\end{eqnarray}
the updating rule of  Eq.~(\ref{eq:linecost}) is simplified as follows.
\begin{align}
L_\mathbf{r}(\mathbf{p}, d) &= C_\mathbf{p}(d) + \min_{d' \in D} \left[ \bar{L}_\mathbf{r}(\mathbf{p} - \mathbf{r}, d') + V_\mathbf{pq}(d, d') \right]\\
&= C_\mathbf{p}(d) + \min \{ \bar{L}_\mathbf{r}(\mathbf{p} - \mathbf{r}, d),
\bar{L}_\mathbf{r}(\mathbf{p} - \mathbf{r}, d-1) + P_1,
\bar{L}_\mathbf{r}(\mathbf{p} - \mathbf{r}, d+1) + P_1,
P_2
\}
 \label{eq:linecost_impl}
\end{align}
Then, the scan-line costs by the 8 directions are aggregated as
\begin{eqnarray}
S(\mathbf{p}, d) = \sum_{\mathbf{r}} L_\mathbf{r}(\mathbf{p}, d),
\end{eqnarray}
from which the disparity estimate at each pixel $\myp$ is retrieved as
\begin{eqnarray}
\myD_\myp = \argmin_{d \in D} S(\mathbf{p}, d). \label{eq:aggcost}
\end{eqnarray}
Recently, Drory~\etal~\cite{Drory2014} showed that the SGM algorithm is a variant of message passing algorithms such as belief propagation and TRW-T~\cite{Wainwright2002} that approximately optimize Eq.~(\ref{eq:stereo}).
Here, the coefficient $c$ in Eq.~(\ref{eq:stereo}) is a scaling factor that accounts for an overweighting effect on the data term during SGM ($c = 1/8$ when using 8 directions)~\cite{Drory2014}.

Drory~\etal~\cite{Drory2014} also proposed an uncertainty measure $\myU$ that is computed as 
\begin{eqnarray}
\myU(\myp) = \min_{d} \sum_{\mathbf{r}} L_\mathbf{r}(\mathbf{p}, d)  - \sum_{\mathbf{r}}  \min_{d} L_\mathbf{r}(\mathbf{p}, d). \label{eq:uncert}
\end{eqnarray}
$\myU(\myp)$ is lower-bounded by 0, and becomes 0 when minimizers of  8 individual scan-line costs agree.
Since the first and second term in Eq.~(\ref{eq:uncert}) are respectively computed in Eqs.~(\ref{eq:aggcost}) and ~(\ref{eq:linecost}), the computation of $\myU(\myp)$ essentially does not require computational overhead.

In our implementation of SGM, we use the data term $C_\myp(\myD_\myp)$ defined using truncated normalized cross-correlation in Eq.~(\ref{eq:tncc}) in the main paper.
The smoothness penalties $P_1$ and $P_2$ are defined as follows.
\begin{equation}
P_1 = \lambda_\text{sgm} / |\myp - \mathbf{q}|
\end{equation}
\begin{equation}
P_2 = P_1 \left(\beta + \gamma w_{\myp\myq}^\text{col} \right)
\end{equation}
Here, $w_{\myp\myq}^\text{col}$ is the color edge-based weight used in Eq.~(\ref{eq:potts}) and 
we use parameters of $(\lambda_\text{sgm}, \beta, \gamma) = (200/255, 2, 2)$.
The disparity range is fixed as $\{D_\text{min}, \cdots, D_\text{max}\}= \{0, \cdots, 255\}$ for the original image size of KITTI (since we downscale the images by a factor of 0.65, the disparity range is also downscaled accordingly).
We also set the confidence threshold $\tau_\text{u}$ for the uncertainty map $\myU$ to 2000 by visually inspecting $\myU(\myp)$. 

\section{SGM Flow}
We have extended the SGM algorithm for our optical flow problem in Sec.~\ref{sec:optflow}.
Here, we estimate the flow map
$\myF_\myp = \myF(\myp) : \Omega \to R$ (where $R=([u_\text{min}, u_\text{max}] \times [v_\text{min}, v_\text{max}])$ is the 2D flow range) by minimizing the following 2D MRF energy.
\begin{equation}
E_{\text{flow}}(\myF) = \sum_{\myp \in \Omega} C'_\myp(\myF_\myp) +  \sum_{(\mathbf{p,q}) \in N} c V'_\mathbf{pq}(\myF_\myp, \myF_\mathbf{q}). \label{eq:flow}
\end{equation}
Similarly to SGM stereo, we use the NCC-based matching cost of Eq.~(\ref{eq:tncc}) for  the data term $C'_\myp(\myF_\myp)$ to evaluate matching photo-consistencies between $I^0_t$ and $I^0_{t+1}$.
We also define the smoothness term as
\begin{eqnarray}
V'_\mathbf{pq}(\myF_\myp, \myF_\myq)=\left\{ \begin{array}{ll}
0 & \text{if} \;\; \myF_\myp = \myF_\myq \\
P_1 & \text{if} \;\; 0 < \|\myF_\myp - \myF_\myq\| \le \sqrt{2} \\
P_2 & \text{otherwise} \\
\end{array} \right.. \label{eq:flowsmooth}
\end{eqnarray}
Since we use integer flow labels, the second condition in Eq.~(\ref{eq:flowsmooth}) is equivalent to saying that the components of the 2D vectors $\myF_\myq = (u_\myq, v_\myq)$ and $\myF_\myp = (u_\myp, v_\myp)$ can at-most differ by 1.
We use the same smoothness penalties $\{P_1, P_2\}$  and the parameter settings with SGM stereo.

The optimization of Eq.~(\ref{eq:flow}) is essentially the same with SGM stereo, but the implementation of updating scan-line costs in Eq.~(\ref{eq:linecost}) was extended to handle the new definition of the pairwise term $V'_\mathbf{pq}$. Therefore, Eq.~(\ref{eq:linecost_impl}) is modified using a flow label $\mathbf{u} = (u, v) \in R$ as follows.
\begin{align}
L_\mathbf{r}(\mathbf{p}, \mathbf{u}) = C_\mathbf{p}(\mathbf{u}) + \min \{ \bar{L}_\mathbf{r}(\mathbf{p} - \mathbf{r}, \mathbf{u}),
\bar{L}_\mathbf{r}(\mathbf{p} - \mathbf{r}, \mathbf{u} + \Delta_{\pm1}) + P_1,
P_2
\}
 \label{eq:linecost_impl_flow}
\end{align}
Here, $(\mathbf{u} + \Delta_{\pm1})$ is enumeration of 8 labels neighboring to $\mathbf{u}$ in the 2D flow space.

\section{Refinement of Flow Maps}
In the optical flow stage of Sec.~\ref{sec:optflow},
we refine flow maps using consistency check and weighted median filtering. Similar schemes are commonly employed in stereo and optical flow methods such as \cite{Zhang2014,Hosni2013,Bleyer2011}. Below we explain these steps.

We first estimate the forward flow map $\myF^0$ (from $I^0_{t}$ to $I^0_{t+1}$) by SGM for only the foreground pixels of the initial segmentation $\tilde{\myS}$ such as shown in Fig.~\ref{fig:postproc}~(a).
Then, using this flow $\myF^0$ and the mask $\tilde{\myS}$, we compute a mask in the next image $I^0_{t+1}$ and estimate the backward flow map  $\myF^1$ (from $I^0_{t+1}$ to $I^0_{t}$) for those foreground pixels.
This produces a flow map such as shown in Fig.~\ref{fig:postproc}~(b).
We filter out outliers in $\myF^0$ using bi-directional consistency check between $\myF^0$ and $\myF^1$ to obtain a  flow map with holes (Fig.~\ref{fig:postproc}~(c)),
whose background is further filled by the rigid flow $\myF_\text{rig}$ (see Fig.~\ref{fig:postproc}~(d)).
Finally, weighted median filtering is applied for the hole pixels followed by median filtering for all foreground pixels to obtain the non-rigid flow estimate such as shown in Fig.~\ref{fig:postproc}~(e).

\begin{figure}[b]
\small
\includegraphics[width=0.33\linewidth]{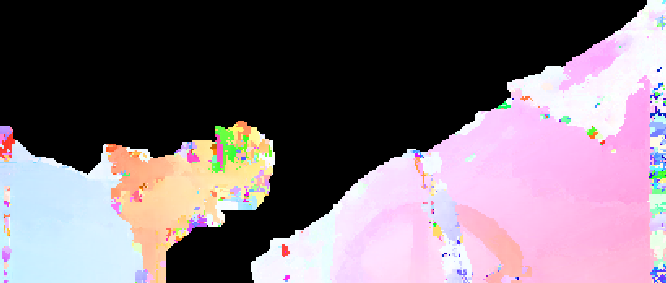}\hfill
\includegraphics[width=0.33\linewidth]{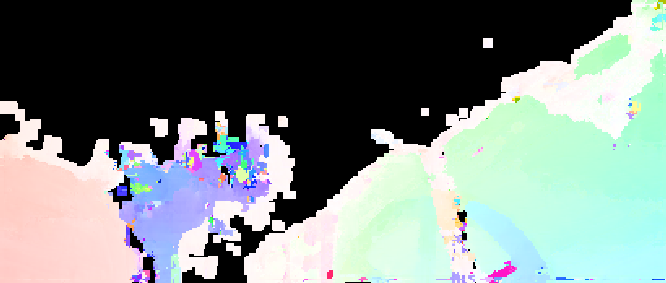}\hfill
\includegraphics[width=0.33\linewidth]{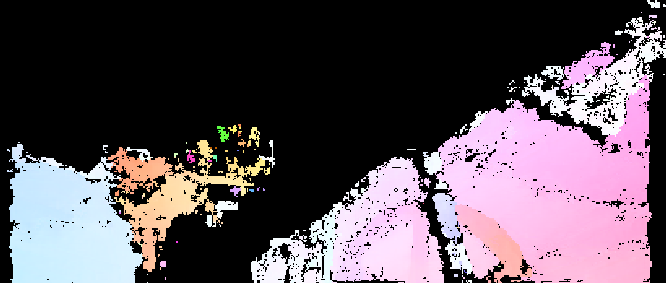}\\
\makebox[0.33\linewidth]{(a) Forward flow map}\hfill
\makebox[0.33\linewidth]{(b) Backward flow map}\hfill
\makebox[0.33\linewidth]{(c) Consistency check}
\vskip 1mm
\includegraphics[width=0.33\linewidth]{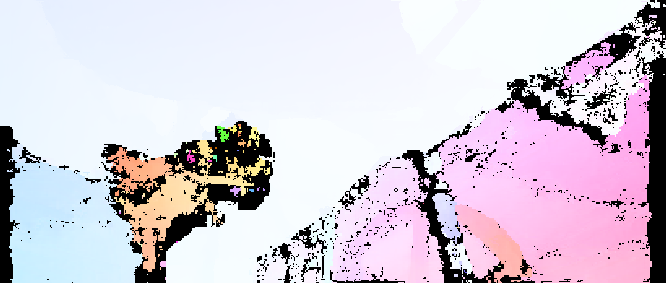}\hfill
\includegraphics[width=0.33\linewidth]{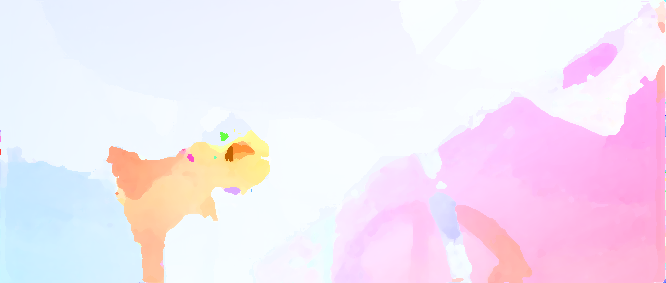}\hfill
\includegraphics[width=0.33\linewidth]{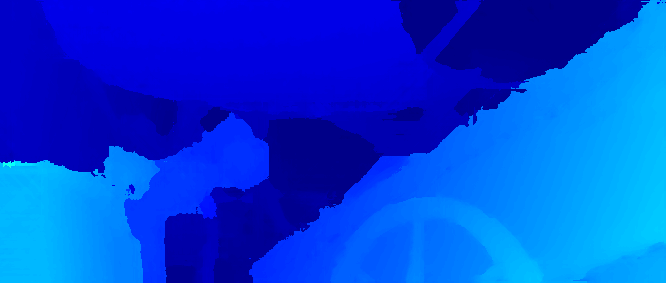}\\
\makebox[0.33\linewidth]{(d) Background filling}\hfill
\makebox[0.33\linewidth]{(e) Weighted median filtering}\hfill
\makebox[0.33\linewidth]{(f) Disparity map as guidance image}
\vskip 0.5mm
\caption{Process of flow map refinement.
}
\label{fig:postproc}
\end{figure}

At the final weighted median filtering step, the filter kernel $\omega^\text{geo}_{\myp\myq} = e^{-d_{\myp\myq}/\kappa_\text{geo}}$ is computed using geodesic distance $d_{\myp\myq}$ on the disparity map $\myD$ (Fig.~\ref{fig:postproc}~(f)) as the guidance image.
For this, we define the distance between two adjacent pixels as
\begin{equation}
\text{dist}(\myp_1, \myp_2) = |\myD(\myp_1) - \myD(\myp_2)| + \|\myp_1 - \myp_2 \| / 100.
\end{equation}
The geodesic distance $d_{\myp\myq}$ is then computed for the pixels in the filter window $\myq \in W_\myp$ as the cumulative shortest-path distance  from $\myq$ to the center pixel $\myp$.
This is efficiently computed using an approximate algorithm~\cite{Toivanen1996}.
 We use the filter window $W_\myp$ of $31 \times 31$ size and $\kappa_\text{geo} = 2$. The subsequent (constant-weight) median filtering further reduces outliers~\cite{Sun2010}, for which we use the window of $5 \times 5$ size.

\section{Segmentation Ground Prior}
The segmentation ground prior term mentioned in Sec.~\ref{sec:details} is computed as follows.
First, we detect the ground plane from the disparity map $\myD(\myp)$.
We use RANSAC to fit a disparity plane $[d = au + bv + c]$ defined on the 2D image coordinates.
Here, we assume that the cameras in the stereo rig are upright. Therefore, during RANSAC we choose  disparity planes whose $b$ is positive and high and $|a|$ is relatively small.
Then, we compute the disparity residuals between $\myD$ and the ground plane as $r_\myp = |\myD_\myp - (a \myp_u + b\myp_v + c)|$, where  $(a, b, c)$ are the obtained  plane parameters.
Our ground prior as a cue of background is then defined as follows.
\begin{equation}
C_\myp^\text{gro} = \lambda_\text{gro} \Big( \min(r_\myp, \tau_\text{gro}) / \tau_\text{gro} -1\Big)
\end{equation}
When $r_\myp = 0$, $C_\myp^\text{gro}$ strongly favors  background, and when $r_\myp$ increases to  $\tau_\text{gro}$, it becomes 0.
The thresholding value $\tau_\myp$ is set to  $0.01 \times D_\text{max}$. We use $\lambda_\text{gro} = 10$.

\begin{figure}[t]
\small
\includegraphics[width=0.485\linewidth]{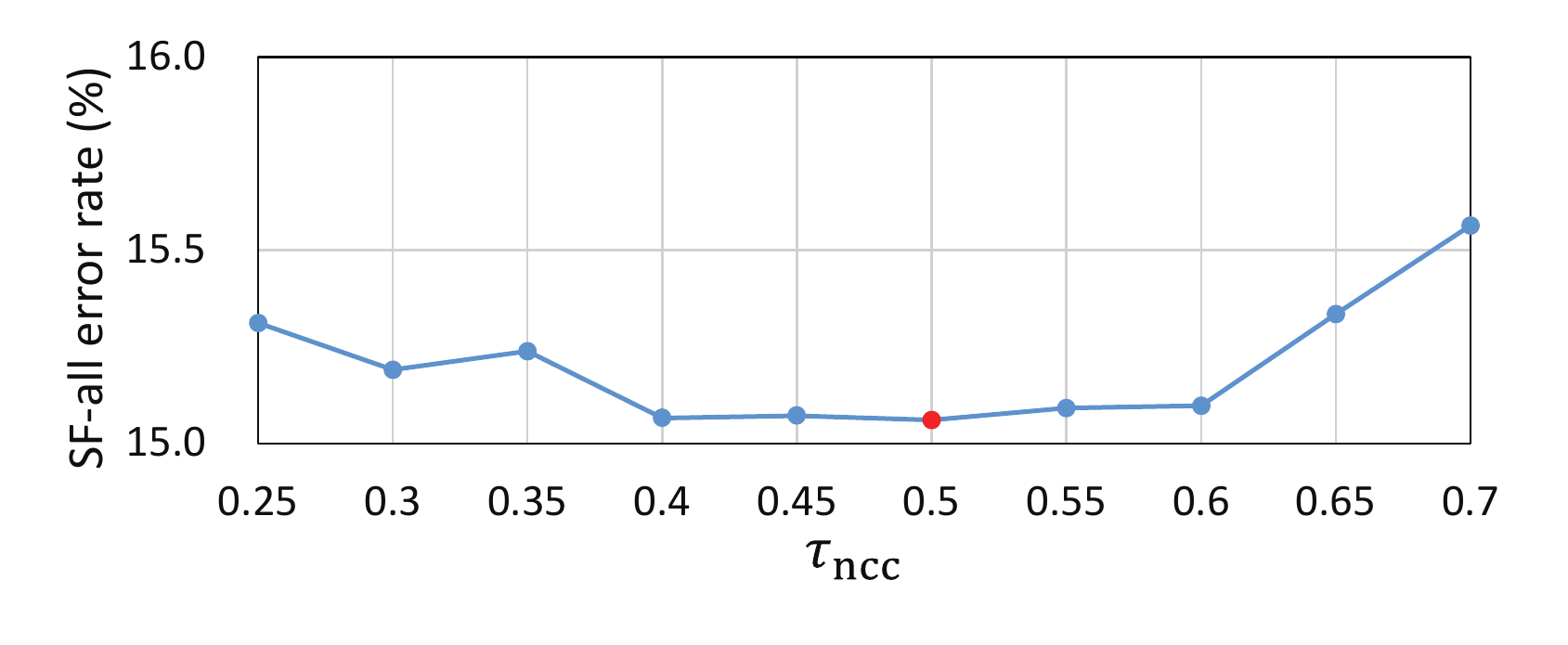}\hfil
\includegraphics[width=0.485\linewidth]{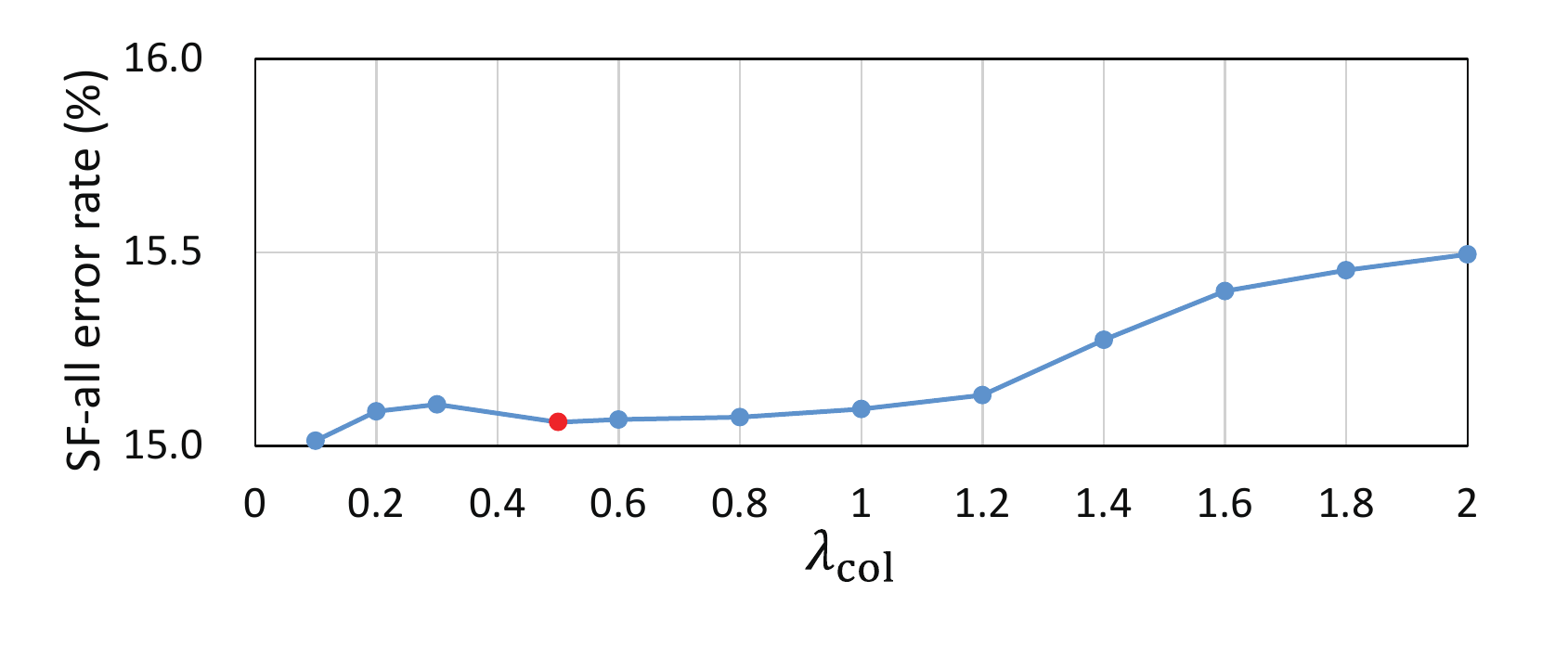}
\caption{Profiles of scene flow accuracies with reference to parameters $\tau_\text{ncc}$ (left) and $\lambda_\text{col}$ (right). The error rates are evaluated on 200 training sequences from KITTI.
The scores with the default parameter settings are colored by red.
}
\label{fig:plot_params}
\end{figure}

\section{Parameter Settings}
In this section, we explain our strategy of tuning parameters and also show effects of some parameters.
Most of the parameters  can be easily interpreted and tuned, and our method is fairly insensitive to parameter settings.

For example, the effects of the threshold $\tau_\text{u}$ for the uncertainty map $\myU$ (Sec.~\ref{sec:stereo}), the threshold $\tau_{w}$ for the patch-variance weight $\omega_\myp^\text{var}$ (Sec.~\ref{sec:segment}), and $\kappa_3$ of the image edge-based weight $\omega_{\myp\myq}^\text{str}$ (Sec.~\ref{sec:segment}) can be easily analyzed by direct visualization as shown in Figure~\ref{fig:stereo}~(b), Figures~\ref{fig:segment}~(b) and (e).

The parameters of SGM (discussed in Sec.~\ref{sec:supp_sgm_stereo}) can be tuned independently from the whole algorithm.

For the weights $(\lambda_\text{ncc}, \lambda_\text{flo}, \lambda_\text{col}, \lambda_\text{potts})$ in Sec.~\ref{sec:segment}, we first tuned $(\lambda_\text{ncc}, \lambda_\text{flo}, \lambda_\text{potts})$ on a small number of sequences.
Since the ranges of the NCC appearance term~(Eq.~(\ref{eq:seg_ncc})) and flow term (Eq.~(\ref{eq:seg_flo})) are limited to $[-1, 1]$, they are easy to interpret.
Then, we tuned $\lambda_\text{col}$ of the color term (Eq.~(\ref{eq:color})).
Here, $\lambda_\text{potts} / \lambda_\text{col}$ is known to be usually around 10 - 60 from previous work~\cite{Rother2004,Boykov2001}.

Even though we fine-tuned $\tau_\text{ncc}$ and $\lambda_\text{col}$ for Sintel, they are insensitive on KITTI image sequences. We show the effects of these two parameters for KITTI training sequences in Figure~\ref{fig:plot_params}.
The threshold $\tau_\text{ncc}$ for NCC-based matching costs was adjusted for Sintel because its synthesized images have lesser image noise compared to real images of KITTI.
Also, the weight $\lambda_\text{col}$ was adjusted for Sintel, to increase the weight on the prior color term~(Sec.~\ref{sec:details}).
For Sintel sequences, sometimes moving objects stop moving on a few frames and become stationary momentarily. In such cases, increasing $\lambda_\text{col}$ improves the temporal coherence of the motion segmentation results.
In the future we will improve the scheme for online learning of the prior color models, which will improve temporal consistency of motion segmentation and also will make $\lambda_\text{col}$ more insensitive to settings.



{\small
\bibliographystyle{ieee}
\bibliography{references}
}

\else

\fi

\end{document}